\documentclass[10pt,journal,compsoc]{IEEEtran}

\ifCLASSOPTIONcompsoc
\else
  \usepackage{cite}
\fi

\ifCLASSINFOpdf
\else
\fi

\usepackage{mdwmath}
\usepackage{mdwtab}

\usepackage{eqparbox}

\ifCLASSOPTIONcompsoc
 \usepackage[caption=false,font=footnotesize,labelfont=sf,textfont=sf]{subfig}
\else
 \usepackage[caption=false,font=footnotesize]{subfig}
\fi

\usepackage{stfloats}

\ifCLASSOPTIONcaptionsoff
 \usepackage[nomarkers]{endfloat}
\let\MYoriglatexcaption\caption
\renewcommand{\caption}[2][\relax]{\MYoriglatexcaption[#2]{#2}}
\fi

\usepackage{url}

\usepackage[numbers,sort]{natbib}
\usepackage{hyperref}

\usepackage{xspace}
\usepackage{bbm}
\input{mathlig}
\usepackage{mathtools}
\usepackage{relsize}
\usepackage{mathrsfs}
\usepackage{dsfont}
\DeclarePairedDelimiterX{\inp}[2]{\langle}{\rangle}{#1, #2}
\makeatletter
\newcommand*\bigcdot{\mathpalette\bigcdot@{.5}}
\newcommand*\bigcdot@[2]{\mathbin{\vcenter{\hbox{\scalebox{#2}{$\m@th#1\bullet$}}}}}
\makeatother

\newcommand{\muspace}{\mspace{1mu}}

\DeclareRobustCommand{\scond}{\mathchoice{\muspace\vert\muspace}{\vert}{\vert}{\vert}}
\mathlig{|}{\scond}

\DeclareRobustCommand{\discint}{\mathchoice{\mspace{-1.5mu}:\mspace{-1.5mu}}{\mspace{-1.5mu}:\mspace{-1.5mu}}{:}{:}}
\mathlig{::}{\discint}
\newcommand{\suchthat}{\mathchoice{\colon}{\colon}{:\mspace{1mu}}{:}}

\def\var{\mathop{\rm Var}\nolimits}%

\newcommand{\Dc}{\mathcal{D}}

\newcommand{\Mc}{\mathcal{M}}

\newcommand{\Rc}{\mathcal{R}}

\newcommand{\Xc}{\mathcal{X}}
\newcommand{\Yc}{\mathcal{Y}}

\newcommand{\Xv}{{\bf X}}
\newcommand{\Yv}{{\bf Y}}
\newcommand{\Zv}{{\bf Z}}
\newcommand{\Uv}{{\bf U}}

\newcommand{\Vv}{{\bf V}}

\newcommand{\xv}{{\bf x}}
\newcommand{\yv}{{\bf y}}
\newcommand{\zv}{{\bf z}}
\newcommand{\uv}{{\bf u}}
\newcommand{\vv}{{\bf v}}
\newcommand{\sv}{{\bf s}}

\newcommand{\ub}{{\mathbf u}}
\newcommand{\vb}{{\mathbf v}}

\newcommand{\xb}{{\mathbf x}}
\newcommand{\Xb}{{\mathbf X}}
\newcommand{\yb}{{\mathbf y}}
\newcommand{\Yb}{\mathbf{Y}}
\newcommand{\zb}{{\mathbf z}}

\def\a{\alpha}
\def\b{\beta}

\def\d{\delta}

\def\eps{\epsilon}
\def\th{\theta}

\DeclareMathOperator\E{\textsf{E}}

\newcommand\eg{e.g.,\xspace}
\newcommand\ie{i.e.,\xspace}
\def\textiid{i.i.d.\@\xspace}
\newcommand\iid{\ifmmode\text{ i.i.d. } \else \textiid \fi}

\newcommand{\Real}{\mathbb{R}}

\def\mathllap{\mathpalette\mathllapinternal}
\def\mathllapinternal#1#2{%
  \llap{$\mathsurround=0pt#1{#2}$}}

\def\clap#1{\hbox to 0pt{\hss#1\hss}}
\def\mathclap{\mathpalette\mathclapinternal}
\def\mathclapinternal#1#2{%
  \clap{$\mathsurround=0pt#1{#2}$}}

\let\oldstackrel\stackrel
\renewcommand{\stackrel}[2]{\oldstackrel{\mathclap{#1}}{#2}}

\DeclarePairedDelimiterX{\infdivx}[2]{(}{)}{%
  #1~\|~#2%
}

\renewcommand{\hbar}{h\mathllap{\overline{\vphantom{h}\hphantom{\rule{4.6pt}{0pt}}}\mspace{0.77mu}}}

\catcode`~=11 %
\newcommand{\urltilde}{\kern -.06em\lower -.06em\hbox{~}\kern .02em}
\catcode`~=13 %

\DeclarePairedDelimiter{\norm}{\lVert}{\rVert}
\DeclarePairedDelimiter{\abs}{\lvert}{\rvert}
\usepackage{xparse}

\makeatletter
\let\oldabs\abs
\def\abs{\@ifstar{\oldabs}{\oldabs*}}
\let\oldnorm\norm
\def\norm{\@ifstar{\oldnorm}{\oldnorm*}}
\let\oldset\set
\def\set{\@ifstar{\oldset}{\oldset*}}
\makeatother

\newcommand{\defeq}{\mathrel{\mathop{:}}=}
\newcommand{\eqdef}{\mathrel{\mathop{=}}:}

\usepackage{subfig}
\usepackage{booktabs}       %
\usepackage{svg}
\usepackage{amsmath,amssymb,amsfonts}
\usepackage{nicefrac}       %
\usepackage{makecell}
\usepackage{wrapfig}
\usepackage{float}
\usepackage{adjustbox}
\usepackage[normalem]{ulem}

\usepackage{mathtools}
\usepackage{empheq}
\usepackage{xcolor}
\makeatletter
\newcommand{\ccell}[3][]{%
  \kern-\fboxsep
  \if\relax\detokenize{#1}\relax
    \expandafter\@firstoftwo
  \else
    \expandafter\@secondoftwo
  \fi
  {\colorbox{#2}}%
  {\colorbox[#1]{#2}}%
  {#3}\kern-\fboxsep
}
\makeatother
\definecolor{cellgray}{gray}{0.9}
\AtBeginDocument{\setlength{\cmidrulekern}{0.3em}}

\allowdisplaybreaks

\renewcommand{\E}{\mathbb{E}}
\newcommand*\diff{\mathop{}\!\mathrm{d}}

\let\oldpartial\partial
\renewcommand*{\partial}{\mathop{}\!\oldpartial}
\DeclareMathOperator*{\minimize}{\mathrm{minimize}~}

\def\xb{{\mathbf x}}
\def\yb{{\mathbf y}}
\def\zb{{\mathbf z}}
\def\Zb{\mathbf Z}

\def\Vb{\mathbf V}
\def\Wb{\mathbf W}

\usepackage{amsmath,amssymb,amsthm}
\newtheorem{theorem}{Theorem}

\newtheorem{proposition}[theorem]{Proposition}
\theoremstyle{definition}

\newtheorem{remark}[theorem]{Remark}

\usepackage{soul}

\newcommand{\Dkl}{D_{\mathsf{KL}}\infdivx*}
\newcommand{\Dsym}{D_{\mathsf{sym}}}
\let\oldpartial\partial
\renewcommand*{\partial}{\mathop{}\!\oldpartial}

\newcommand{\model}{{\mathsf{model}}}
\newcommand{\joint}{{\mathsf{\to xy}}}

\renewcommand{\var}{{\mathsf{xy\to}}}

\newcommand{\ytox}{{\mathsf{y\to x}}}
\newcommand{\xtoy}{{\mathsf{x\to y}}}
\newcommand{\xtoytox}{\mathsf{x\leftrightarrow y}}
\newcommand{\qvar}{q_{\var}}
\newcommand{\pmodel}{p_{\model}}
\newcommand{\pjoint}{p_{\joint}}

\newcommand{\pytox}{p_{\ytox}}
\newcommand{\pxtoy}{p_{\xtoy}}
\newcommand{\xyzuv}{\mathsf{xyzuv}}

\newcommand{\qdata}{q_{\mathsf{data}}}

\newcommand{\wv}{\mathbf{w}}

\newcommand\numberthis{\addtocounter{equation}{1}\tag{\theequation}}

\begin{document}
\title{Learning with Succinct Common Representation Based on Wyner's Common Information}

\author{J.~Jon~Ryu,~\IEEEmembership{Student Member,~IEEE,}
        Yoojin~Choi,~\IEEEmembership{Member,~IEEE,}
        Young-Han~Kim,~\IEEEmembership{Fellow,~IEEE}\\
        Mostafa El-Khamy,~\IEEEmembership{Senior Member,~IEEE}
        Jungwon Lee,~\IEEEmembership{Fellow,~IEEE}%
\IEEEcompsocitemizethanks{
\IEEEcompsocthanksitem J.~J.~Ryu is
with the Department of Electrical and Computer Engineering, University of California San Diego, La Jolla, CA 92093, USA.\protect\\
E-mail: \href{mailto:jongha.ryu@gmail.com}{jongha.ryu@gmail.com}
\IEEEcompsocthanksitem Y.~Choi is with SOC R\&D, Samsung Semiconductor Inc., San Diego, CA 92121, USA.\protect\\
E-mail: \href{mailto:yoojin.c@samsung.com}{yoojin.c@samsung.com}
\IEEEcompsocthanksitem Y.-H.~Kim is with the Department of Electrical and Computer Engineering, University of California San Diego, La Jolla, CA 92093, USA and Gauss Labs Inc, Palo Alto, CA 94301, USA.\protect\\
E-mail: \href{mailto:yhk@ucsd.edu}{yhk@ucsd.edu}
\IEEEcompsocthanksitem M.~El-Khamy is with SOC R\&D, Samsung Semiconductor Inc., San Diego, CA 92121, USA and Alexandria University, Alexandria Governorate 5424041, Egypt.\protect\\
E-mail: \href{mailto:mostafa.e@samsung.com}{mostafa.e@samsung.com}
\IEEEcompsocthanksitem J.~Lee is with SOC Development Office, Samsung Electronics, Gyeonggi-do 18448, Republic of Korea.\protect\\
E-mail: \href{mailto:jungwon2.lee@samsung.com}{jungwon2.lee@samsung.com}
}%
}

\IEEEtitleabstractindextext{%
\begin{abstract}
A new bimodal generative model is proposed for generating conditional and joint samples, accompanied with a training method with learning a succinct bottleneck representation.
The proposed model, dubbed as the variational Wyner model, is designed based on two classical problems in network information theory---distributed simulation and channel synthesis---in which Wyner's common information arises as the fundamental limit on the succinctness of the common representation.
The model is trained by minimizing the symmetric Kullback--Leibler divergence between variational and model distributions with regularization terms for common information, reconstruction consistency, and latent space matching terms, which is carried out via an adversarial density ratio estimation technique.
The utility of the proposed approach is demonstrated through experiments for joint and conditional generation with synthetic and real-world datasets, as well as a challenging zero-shot image retrieval task.
\end{abstract}

}

\maketitle

\IEEEdisplaynontitleabstractindextext

\IEEEpeerreviewmaketitle

\section{Introduction}
Over the last decade, we have witnessed an uncountable number of successes of deep learning, which are mostly attributed to its power of learning a good, low-dimensional representation of data~\citep{Bengio--Courville--Vincent2013}. 
The importance of representation learning has become more significant than ever in the last few years, as represented by a recent paradigm shift towards a task-agnostic learning framework~\citep{Bommasani--EtAl2021Foundation} and the emerging successes in self-supervised learning~\citep{Chen--Kornblith--Norouzi--Hinton2020}.
Despite the practical breakthroughs, however, answers to the fundamental questions like ``what is a good representation?'' and ``how can we find such a representation?'' are still unsatisfactory.

\begin{figure}[ht]
    \centering
    \includegraphics[width=.3\textwidth]{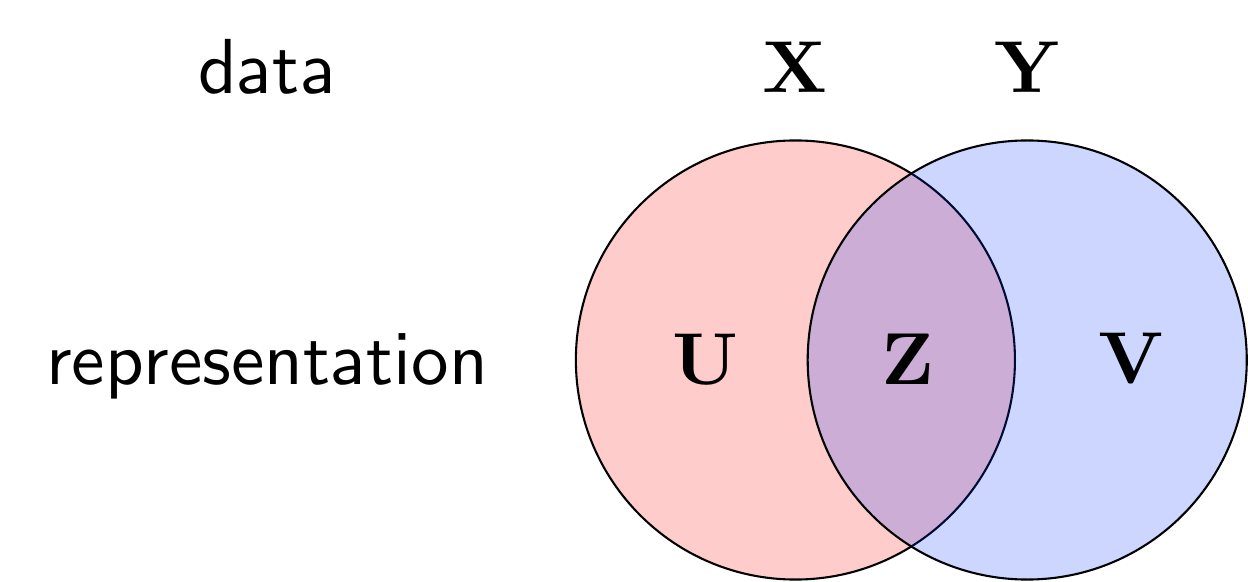}
    \caption{A Venn-diagram schematic for cross-domain disentanglement.}
    \label{fig:venn}
\end{figure}

In this context, we study how to learn a good joint representation of a pair of random vectors $(\Xv,\Yv)$ with complex dependence from data, with the following structure:
we wish to learn a \emph{structured} representation $(\Zv,\Uv,\Vv)$ of $(\Xv,\Yv)$ such that $(\Zv,\Uv)$ and $(\Zv,\Vv)$ capture the information of $\Xv$ and $\Yv$, respectively.
Here, $\Zv$ captures the commonality of $(\Xv,\Yv)$, which we thus call a \emph{common representation} of $(\Xv,\Yv)$;
$\Uv$ and $\Vv$, which we call \emph{local representations}, correspond to the remaining information on
$\Xv$ and $\Yv$, respectively.
See Fig.~\ref{fig:venn} for a Venn-diagram schematic of the structured representation.
This problem is often referred to as the \emph{cross-domain disentanglement} problem~\citep{Gonzalez-Garcia--VanDeWeijer--Bengio2018} in the machine learning literature and has numerous applications including joint and conditional generative tasks (also known as domain transfer or image-to-image translation) and cross-domain retrieval tasks~\citep{Zhu--Zhang--Pathak--Darrell--Efros--Wang--Shechtman2017,Gonzalez-Garcia--VanDeWeijer--Bengio2018,Lee--Tseng--Jia-Bin--Singh--Yang2018,Liu--Liu--Yeh--Wang2018,Huang--Liu--Belongie--Kautz2018,Yu--Chen--Li--Liu--Li2019,Press--Galanti--Benaim--Wolf2020}.

The main difficulty in learning such joint distributions with disentangled representations is that there have been no proper criterion for cross-domain disentanglement that defines an \emph{optimal} common representation.
Indeed, existing approaches which are mostly from the deep learning literature focus on developing a network architecture and/or a set of ad-hoc loss functions that promote the degree of disentanglement, not defining an optimal common representation of a joint distribution.
Even a few existing information-theoretic proposals on learning a good bottleneck representation such as the famous information bottleneck principle~\citep{Tishby--Pereira--Bialek1999} and a recent proposal~\citep{Hwang--Kim--Hong--Kim2020IIAE} based on interactive information~\citep{McGill1954} do not define what an optimal representation is and what they aim to look for.

Observe that there are two bad extremes for the common representation.
On one hand, we can use raw data as the common representation $\Zv=\Xv$ or $\Zv=\Yv$, which contain maximal information of the pair, but may not be helpful from the view of a user who wishes to perform a downstream task based on it, as there exists a large degree of \emph{redundancy}. 
On the other hand, one may choose a common representation $\Zv$ as a constant; albeit being the simplest, it discards essential information about the pair and is thus not useful representation. %
Hence, it is natural to assume that an optimal representation must lie somewhere in between, \ie capturing the most \emph{succinct} possible representation, as well as maintaining all the commonality of the pair.

In this paper, as a first proposal to the missing definition of an optimal common representation,
we propose a new representation learning principle inspired by network information theory. 
To motivate our perspective, consider the following game between Alice (``encoder'') and Bob (``decoder'') that captures the problem setting of \emph{conditional generation}.
Given an image of a child's photo $\Xv$, Alice is asked to encode $\Xv$ and send its description $\Zv$ to Bob who draws a portrait $\Yv$ of how the child will grow up based on it.
In this game, Bob wishes to draw nice adulthood portraits, as various as possible, given a child's photo.
In this cooperative game, Alice needs to help Bob in the process by providing a \emph{good} description $\Zv$ of the child's photo $\Xv$. 
Intuitively, seeking the \emph{most succinct description} $\Zv$ that contains information \emph{common in $\Xv$ and $\Yv$} may be beneficial in their guessing processes, since Alice need not describe any extra information beyond that is contained in $\Xv$ and Bob need not filter out any redundant information from $\Zv$.

\begin{figure*}[!tbp]%
\centering
\subfloat[Channel synthesis]
{\includegraphics[scale=0.7]{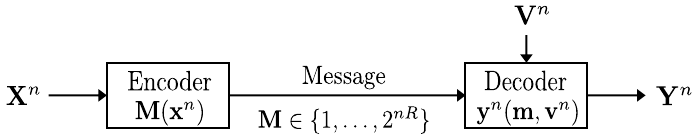}}
\hfill
\subfloat[Single-letter characterization of (a)]
{\includegraphics[scale=0.7]{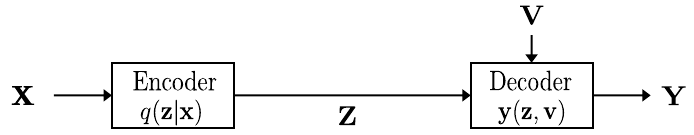}}\\
\subfloat[Distributed simulation]
{\includegraphics[scale=0.7]{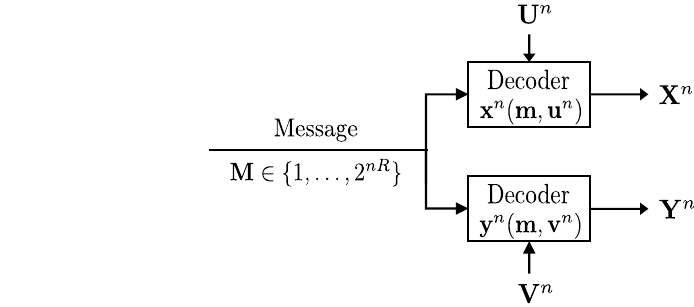}}
\hfill
\subfloat[Single-letter characterization of (c)]
{\includegraphics[scale=0.7]{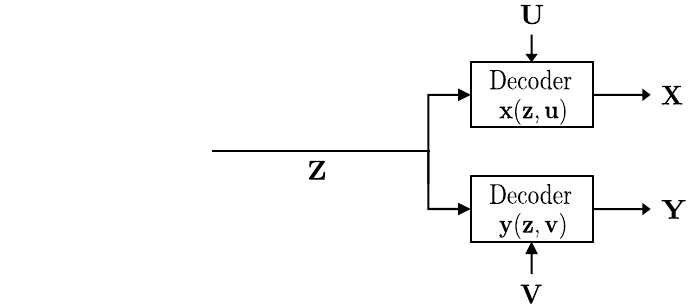}}
\caption{Schematics for channel synthesis from $\Xv$ to $\Yv$ (a,b), and distributed simulation of $(\Xv,\Yv)$ (c,d).
(a,c) and (b,d) correspond to the operational definition and the single-letter characterization of each problem, respectively. The local randomness $\Uv$ and $\Vv$ make the decoders stochastic.}
\label{fig:schematics}
\end{figure*}

P.~Cuff~(\citeyear{Cuff2013}) formulated this game of conditional generation as the \emph{channel synthesis} problem in network information theory depicted in Fig.~\ref{fig:schematics}.
Given a joint distribution $\qdata(\xv,\yv)=\qdata(\xv)\qdata(\yv|\xv)$, Alice and Bob want to generate $\Yv$ according to $\qdata(\yv|\xv)$ based on a sample from $\qdata(\xv)$.
In this problem, Alice wishes to find the most succinct description $\Zv$ of $\Xv$ (a child's photo) such that $\Yv$ (her adulthood portrait) can be simulated by Bob according to the desired distribution using this description and local randomness $\Vv$ (new features to draw a portrait of adults that are not contained in photos of children).
The minimum description rate for such conditional generation is characterized by \emph{Wyner's common information (CI)}~\citep{Wyner1975,El-Gamal--Kim2011}, which is denoted by $J(\Xv;\Yv)$ and defined as the optimal value of the following optimization problem, which we will call \emph{Wyner's optimization problem} hereafter:
\begin{empheq}[]{equation}\label{eq:wyner}
\begin{aligned}
\minimize & I(\Xv,\Yv;\Zv) 
\\
\text{~subject to~} &
    (\Xv,\Yv,\Zv)\sim \qdata(\xv,\yv)q_\phi(\zv|\xv,\yv)\\
    & \Xv\leftrightarrow\Zv\leftrightarrow\Yv\\
\text{~variables~} & q_\phi(\zv|\xv,\yv).
\end{aligned}
\end{empheq}
Here, $I(\Xv,\Yv;\Zv)$ is the mutual information between $(\Xv,\Yv)$ and  $\Zv$, and $\Xv\leftrightarrow\Zv\leftrightarrow\Yv$ denotes that $\Xv,\Zv,\Yv$ form a Markov chain, \ie $\Xv$ is independent of $\Yv$ given $\Zv$~\citep{Cover--Thomas2006}.

Notably, the same quantity $J(\Xv;\Yv)$ arises as the fundamental limit of the \emph{distributed simulation} of correlated sources studied originally by A.~Wyner~(\citeyear{Wyner1975}) 
in which two distributed agents wish to simulate a target distribution $\qdata(\xv,\yv)$ based on the least possible amount of shared common randomness; see Fig.~\ref{fig:schematics} (c,d).
As the channel synthesis problem can be viewed as an information-theoretic counterpart of conditional generation, the distributed simulation corresponds to  joint generation.

Thus motivated from these observations, in this paper,
we suggest to define an optimal common representation of a given joint distribution $q(\xb,\yb)$ as the optimal solution of
Wyner's optimization problem~\eqref{eq:wyner}, and use the probabilistic model with a succinct common representation for both joint and conditional generation tasks.

Towards its application in generative modeling, in Section~\ref{sec:models}, we first propose a probabilistic model that finds a common representation, based on the single-letter characterizations of the distributed simulation and channel synthesis problems; see Fig.~\ref{fig:schematics}(b),(d).
Note that the resulting probabilistic models, which we call the \emph{variational Wyner model} as a whole, follow the Markov chain $\Xv\leftrightarrow\Zv\leftrightarrow\Yv$, which also appear in Wyner's optimization problem~\eqref{eq:wyner} as a constraint.
Here, the mutual information $I(\Xv,\Yv;\Zv)$ emerges as a measure of the complexity of the common representation $\Zv$ characterized by $q(\zv|\xv,\yv)$; see Remark~\ref{rem:ci}. 

Now in the learning setting, where we only have access to the joint distribution $q(\xb,\yb)$ via its samples,
we propose to train the probabilistic model based on Wyner's optimization problem~\eqref{eq:wyner}.
We will first derive from \eqref{eq:wyner} a set of distribution matching losses and CI regularization losses as the main objectives.
To learn with samples more effectively, we further propose auxiliary objectives such as reconstruction losses and latent matching losses. 
See Section~\ref{sec:objectives}.

In Section~\ref{sec:training}, we discuss how to train the variational Wyner model based on the proposed training objective.
As an effective training trick, we specifically adopt an approximate training method using an variational density ratio estimation technique~\citep{Pu--Wang--Henao--Chen--Gan--Li--Carin2017SymmetricVAE}.
With this training trick, after all, the proposed generative model can be viewed as an adversarially learned bimodal autoencoder.

Section~\ref{sec:related} discusses related work on Wyner's CI from the information theory literature, existing information-theoretic approaches such as \citep{Tishby--Pereira--Bialek1999} and \citep{Hwang--Kim--Hong--Kim2020IIAE}, and other bimodal generative models and cross-domain disentanglement approaches.

In Section~\ref{sec:exp},
we justify this framework (the model, the training objectives, and the training method as a whole) by empirically showing that learning with its deep generative model manifestation can indeed improve an \emph{empirical quality} in generative tasks and various downstream tasks for synthetic and real-world dataset, demonstrating that the amount of CI captured in $\Zv$ can be controlled to improve the quality of the model.
We defer the details of training schemes and network architectures to Appendix in Supplementary Material.

\section{Probabilistic Models}
\label{sec:models}
In this section, we define all probabilistic model components for joint and conditional sampling tasks based on the Markov chain $\Xv\leftrightarrow\Zv\leftrightarrow\Yv$ and the single-letter characterizations in Fig.~\ref{fig:schematics}~(b,d). 

\subsection{Joint Model}
\label{sec:joint_model}
As a generative model for modeling the joint distribution $\qdata(\xv,\yv)$, we consider the latent variable model $p_\th(\zv)p_\th(\xv|\zv)p_\th(\yv|\zv)$.
Here, $\Zv\sim p_\th(\zv)$ signifies the common randomness fed into the \emph{probabilistic} decoders $p_\th(\xv|\zv)$ and $p_\th(\yv|\zv)$.
We parameterize the probabilistic decoders $p_\th(\xv|\zv)$ and $p_\th(\yv|\zv)$ by (deterministic) functions $\xv_\th(\zv,\uv)$ and $y_\th(\zv,\vv)$ with independent \emph{local randomness} $\Uv\sim p_\th(\uv)$ and $\Vv\sim p_\th(\vv)$, as depicted in the single letter characterization of distributed simulation (Fig.~\ref{fig:schematics}~(d)).
With a slight abuse of notation, we use $\xv_\th(\zv,\uv)$ for a shorthand for the degenerate distribution $\d(\xv-\xv_\th(\zv,\uv))$.
\subsection{Conditional Models}
\label{sec:conditional_models}
To model the conditional distribution $\qdata(\yv|\xv)$, we consider the bottleneck conditional model $q_\th(\zv|\xv)p_\th(\yv|\zv)$ that follows $\Xv\leftrightarrow\Zv\leftrightarrow\Yv$; note that the decoder $p_\th(\yv|\zv)$ is shared by the joint model.
The other direction for modeling $\qdata(\xv|\yv)$ is symmetric and thus omitted.

\subsection{Variational Encoders}
\label{sec:variational_encoders}
In addition to the base components introduced so far from which we can draw joint and conditional samples, we introduce three additional encoders:
\begin{itemize}
\item A joint encoder $q_\phi(\zv|\xv,\yv)$: it plays a key role of an anchor during training, tying the joint and conditional models. 
\item Local encoders $q_\phi(\uv|\zv,\xv)$, $q_\phi(\vv|\zv,\yv)$: these can be viewed as \emph{style extractors} for each modality $\xv$ and $\yv$: if we learn a succinct common representation $q_\phi(\zv|\xv,\yv)$ (e.g., a shared concept) from $(\xv,\yv)$, then $q_\phi(\uv|\zv,\xv)$ captures the remaining randomness $\Uv$ of $\Xv$ (e.g., texture and style). 
\end{itemize}
We will call these encoders \emph{variational} due to a technical reason to be justified when training objectives are introduced below in Section~\ref{sec:objectives_main}.%
These encoders can be used in training by allowing us to enforce the reconstruction consistency of the model as shown in the next section, as well as in several inference tasks such as domain translation; see Remark~\ref{rem:style}.%

\subsection{Variational Wyner Model}
We call the entire model with all the components introduced above, \ie in Sections~\ref{sec:joint_model}--\ref{sec:variational_encoders}, 
as the (bimodal) \emph{variational Wyner model}. 
We remark that in this general framework we may learn multiple models sharing common components, or learn a single model without training the others, depending on the task at hand. 
For example, if one is interested in captioning an image, we may only require learning a conditional model from image to caption, without learning the joint distribution and the conditional distribution of image given caption.
For the cross-domain retrieval task (see Remark~\ref{rem:retrieval} and Section~\ref{sec:zs_sbir}), we require to learn both conditional models.
When multiple models are trained simultaneously, common components such as the joint encoder $q_\phi(\zb|\xb,\yb)$ and local encoders $q_\phi(\ub|\zb,\xb),q_\phi(\vb|\zb,\yb)$ are \emph{shared}. 

\begin{remark}[Conditional independence structure]
\label{rem:conditional_independence}
The components of the variational Wyner model may naturally arise considering the cross-domain disentanglement problem, and indeed similar models have been studied in the literature; see \eg \citep{Gonzalez-Garcia--VanDeWeijer--Bengio2018,Hwang--Kim--Hong--Kim2020IIAE,Wang--Yan--Lee--Livescu2016DeepVCCA}.
The key structural difference of our model compared to existing ones is the conditioning with the common representation $\Zv$ in the local encoders $q_\phi(\uv|\zv,\xv)$, $q_\phi(\vv|\zv,\yv)$, which are designed to satisfy the conditional independence structure implied by the joint model $p_\th(\zv)p_\th(\uv)p_\th(\vv) \xv_\th(\zv,\uv) \yv_\th(\zv,\vv)$, \ie  \[q_\phi(\zv,\uv,\vv|\xv,\yv)=q_\phi(\zv|\xv,\yv)q_\phi(\uv|\zv,\xv)q_\phi(\vv|\zv,\yv),\] 
(see Proposition~\ref{prop:var_encoders} in Appendix), 
while the existing models ignore the conditioning with $\zv$, that is, use variational encoders of the form $q(\uv|\xv)$ and/or $q(\vv|\yv)$~\citep{Gonzalez-Garcia--VanDeWeijer--Bengio2018,Hwang--Kim--Hong--Kim2020IIAE,Wang--Yan--Lee--Livescu2016DeepVCCA}.
In experiments, we empirically validate that this conditioning with $\Zv$ indeed helps learn disentangled representations; see the cross-domain retrieval task experiment in Section~\ref{sec:exp}.
\end{remark}

\begin{figure*}[!htb]
\centering
\centerline{\includegraphics[width=\textwidth]{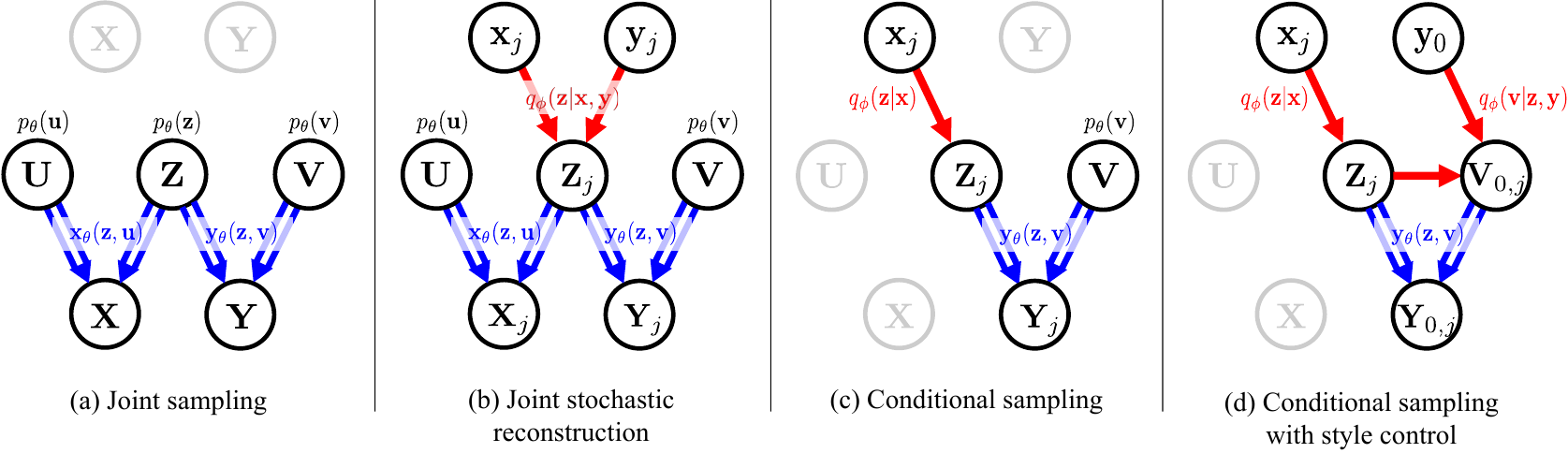}}
\caption{Schematics for selected sampling tasks. Double-line arrows are used to emphasize the deterministic mappings.}
\label{fig:sampling}
\end{figure*}

\begin{remark}[Sampling with style control]\label{rem:style}
Once the Wyner model is properly trained to fit $q(\xb,\yb)$, joint or conditional sampling can be done to simulate sample generation from $q(\xb,\yb)$ or $q(\yb|\xb)$ in a straightforward manner as depicted in Fig.~\ref{fig:sampling}(a,c).
The variational encoders $q_\phi(\ub|\zb,\xb),q_\phi(\vb|\zb,\yb)$ can be used  as a local representation extractor in sampling tasks \emph{with style control}, such as joint stochastic reconstruction~(Fig.~\ref{fig:sampling}(b)) or conditional sampling with style control~(Fig.~\ref{fig:sampling}(d));
here, we elaborate the latter use case. 
Suppose that $(\Xb,\Yb)$ is a pair of correlated images generated from the common concept but from different domains.
Similar to conditional sampling (Fig.~\ref{fig:sampling}(c)), we first draw $\Zv_j$ from $q_\phi(\zv|\xv_j)$. 
Given an image $\yb_0$, we then extract the style information $\Vb_{0,j}$ from  $q_\phi(\vb|\Zv_j,\yb_0)$ (Fig.~\ref{fig:sampling}(d)).
Finally, we generate $\Yv_{0,j}$ from an image $\xb_j$ while replacing the randomly drawn local representation $\Vb\sim p_\th(\vb)$ with the previously extracted style $\Vb_{0,j}$, thereby the generated images $\Yb_{0,j}$ is of the same style as the reference image $\yb_0$.
In a similar manner, we can also perform joint sampling with a fixed style given a style reference data pair $(\xb_0,\yb_0)$, by mixing a randomly drawn common representation $\Zb$ from the prior $p_\th(\zb)$ with the extracted style variables $(\ub_0,\vb_0)$.
\end{remark}

\begin{remark}[Cross-domain retrieval]
\label{rem:retrieval}
Beyond the joint and conditional generation tasks, there is another closely related task which is called the \emph{cross-domain retrieval}~\citep{Gonzalez-Garcia--VanDeWeijer--Bengio2018,Hwang--Kim--Hong--Kim2020IIAE}.
In this task, we are given a reference set $\{\yv_i\}$. 
For a query $\xv_o$, instead of aiming to draw a fresh sample $\yv$ from $q(\yv|\xv_o)$, we wish to \emph{retrieve} relevant $\yv$'s from $\{\yv_i\}$.
We can solve the retrieval task using a trained variational Wyner model over the common representation space, similar to \citep{Gonzalez-Garcia--VanDeWeijer--Bengio2018,Hwang--Kim--Hong--Kim2020IIAE}.
That is, we first find and keep the common representations $\{\zv_i\}$ of reference points $\{\yv_i\}$ using the model encoder $q_\th(\zv|\yv)$. 
Then, given a query $\xv_o$, we find the common representation $\zv_o\sim q_\th(\zv|\xv_o)$ to retrieve the $K$-nearest neighbors of $\zv_o$ from $\{\zv_i\}$ with respect to, say, the cosine similarity. 
We remark that for this task, we only require to learn the conditional models of both directions.
See Section~\ref{sec:zs_sbir} for our experimental results.
\end{remark}

\subsection{Induced Distributions}

The variational Wyner model defines four different distributions over the extended set of variables $(\xv,\yv,\zv,\uv,\vv)$. 
We explicitly write down the distributions below, as we will match the pairs of distributions to train the generative models of our interest in the next section.

The first one is the \emph{variational distribution}, which is defined by the underlying data distribution and the variational encoders:
\begin{align*}
&\qvar(\xv,\yv,\zv,\uv,\vv)\\
&\defeq \qdata(\xv,\yv)q_\phi(\zv|\xv,\yv)q_\phi(\uv|\zv,\xv)q_\phi(\vv|\zv,\yv).
\numberthis
\end{align*}
The other three distributions are the \emph{model} distributions, which correspond to the joint and conditional generative models. 
The joint model induces
\begin{align*}
\pjoint(\xv,\yv,\zv,\uv,\vv)
&\defeq p_\th(\zv)p_\th(\uv)p_\th(\vv)\xv_\th(\zv,\uv)\yv_\th(\zv,\vv),
\numberthis
\end{align*}
and the conditional model that \emph{maps} $\xv$ to $\yv$ induces
\begin{align*}
&\pxtoy(\xv,\yv,\zv,\uv,\vv)\\
&\defeq \qdata(\xv)q_\th(\zv|\xv)q_\phi(\uv|\zv,\xv)p_\th(\vv)\yv_\th(\zv,\vv).
\numberthis
\end{align*}
Symmetrically, the other direction from $\yv$ to $\xv$ induces
\begin{align*}
&\pytox(\xv,\yv,\zv,\uv,\vv)\\
&\defeq \qdata(\yv)q_\th(\zv|\yv)q_\phi(\vv|\zv,\yv)p_\th(\uv)\xv_\th(\zv,\uv).
\numberthis
\end{align*}

Hereafter, for each model $\pmodel\in\{\pjoint, \pxtoy, \pytox\}$ and a subset of variables $\wv\subseteq \{\xv,\yv,\zv,\uv,\vv\}$, we use $\pmodel(\wv)$ to denote the induced distributions over $\wv$.
For example, for $\wv=\{\xv,\yv,\zv\}$, the induced distribution is
\begin{align*}
\pmodel(\xv,\yv,\zv)&\defeq\int\pmodel(\xv,\yv,\zv,\uv,\vv)\diff\uv\diff\vv.
\end{align*}

We remark that while the variational distribution $q_{\var}$ is always consistent with the data distribution but may fail to satisfy $\Xv-\Zv-\Yv$, the model distributions $\pjoint, \pxtoy, \pytox$ may not be consistent with $\qdata(\xv,\yv)$ but always follow the Markov chain.

\begin{remark}[Common information]
\label{rem:ci}
In \eqref{eq:wyner}, we call the mutual information $I_{\var}(\Xv,\Yv;\Zv)$ induced by the variational distribution $q_{\var}(\xv,\yv,\zv)$, \ie
\[
I_{\var}(\Xv,\Yv;\Zv)\defeq \E_{\qdata(\xv,\yv)}[\Dkl{q_\phi(\zv|\xv,\yv)}{\qvar(\zv)}],
\] 
the \emph{variational CI}. Here, $\Dkl{p}{q}$ denotes the Kullback--Leibler (KL) divergence between two distributions $p$ and $q$.
Now, for each model distribution $\pmodel\in\{\pjoint, \pxtoy, \pytox\}$, which is designed to follow $\Xv-\Zv-\Yv$ above, 
we call the the corresponding mutual information between $(\Xv,\Yv)$ and $\Zv$
\begin{align*}
I_{\model}&\defeq I_{\model}(\Xv,\Yv;\Zv)\\
&\defeq \Dkl{\pmodel(\xv,\yv,\zv)}{\pmodel(\xv,\yv)\pmodel(\zv)}
\numberthis
\label{eq:obj_ci_model}
\end{align*}
the \emph{model CI} under $\pmodel$.
\end{remark}

\section{Training Objectives}
\label{sec:objectives}
In this section, we describe a set of training objectives for effectively training the proposed variational Wyner model.

\subsection{Main Objectives} %
\label{sec:objectives_main}
Recall Wyner's optimization problem from the introduction:
\begin{align}
\begin{aligned}
\minimize & I_{\var}(\Xv,\Yv;\Zv) 
\\
\text{~subject to~}
    & \Xv\leftrightarrow\Zv\leftrightarrow\Yv,
\end{aligned}
\label{eq:wyner_recall}
\end{align}
where the variable is the joint encoder $q_{\phi}(\zv|\xv,\yv)$. 
Hence, this optimization problem seeks a joint encoder $q_{\phi}(\zv|\xv,\yv)$ that captures the minimal common information under the Markovity constraint.

For each model distribution $\pmodel\in\{\pjoint, \pxtoy, \pytox\}$, our main learning principle is to train it by seeking a succinct common representation characterized by Wyner's optimization problem, and we reformulate the optimization problem~\eqref{eq:wyner_recall} for each model.
First, since each model distribution $\pmodel(\xv,\yv,\zv)$ follows the Markov chain $\Xv\leftrightarrow\Zv\leftrightarrow\Yv$, 
we replace the Markovity constraint with the following model consistency
\begin{align*}
\pmodel(\xv,\yv,\zv)&\equiv \qdata(\xv,\yv)q_{\phi}(\zv|\xv,\yv),
\end{align*}
\ie the model $\pmodel$ is \emph{consistent} with the target data distribution $\qdata(\xv,\yv)$ and the variational distribution $q_{\phi}(\zv|\xv,\yv)$.
Under this model consistency, we can further replace the variational CI $I_{\var}(\Xv,\Yv;\Zv)$ with the model CI $I_{\model}(\Xv,\Yv;\Zv)$, which is defined in \eqref{eq:obj_ci_model}.
Hence, for each model $\pmodel$, we obtain
\begin{align}
\begin{aligned}
\minimize & I_{\model}(\Xv,\Yv;\Zv) 
\\
\text{~subject to~}
    & \pmodel(\xv,\yv,\zv)\equiv \qdata(\xv,\yv,\zv).
\end{aligned}
\label{eq:wyner_each_model}
\end{align}

The model consistency can be imposed by introducing a distribution matching term
\[\Dc_{\model}^{\mathsf{xyz}}\defeq D(\pmodel(\xv,\yv,\zv),\qdata(\xv,\yv,\zv))\]
and constraining it to be 0 for a choice of positive definite divergence function $D(p,q)$ such as $f$-divergences, Wasserstein distance, or maximum mean discrepancy~\citep{Zhao--Song--Ermon2018}.
Note that each choice of $D(p,q)$ requires different training methods and training procedures.
In this paper, we specifically choose the \emph{symmetric KL divergence}
\[
\Dsym(p(\sv),q(\sv))
\defeq \Dkl{p(\sv)}{q(\sv)}+\Dkl{q(\sv)}{p(\sv)}
\]
which is also known as the \emph{Jeffreys divergence}~\citep{Jeffreys1998}.  \citet{Pu--Wang--Henao--Chen--Gan--Li--Carin2017SymmetricVAE} originally suggested its use in generative modeling as an alternative of the one-sided (reverse) KL diveregence $\Dkl{q}{p}$ of VAEs, 
since it can encourage mode-seeking and mass-covering simultaneously.
Unlike the typical VAE training, however, the symmetric KL divergence necessitates an additional trick to deal with intractable density ratios. We illustrate an approximate training method in Section~\ref{sec:training}.

Therefore, Wyner's optimization problem~\eqref{eq:wyner_each_model} for each model can be written as
\begin{align}
\begin{aligned}
\minimize & I_{\model}(\Xv,\Yv;\Zv) 
\\
\text{~subject to~}
    & \Dc_{\model}^{\mathsf{xyz}}=0.
\end{aligned}
\label{eq:wyner_each_model_2}
\end{align}
We now show that this can be further relaxed as
\begin{align}
& \minimize~ %
\Dc_{\model}^{\xyzuv}+\lambda_{\model}^{\mathsf{CI}} I_{\model}(\Xv,\Yv;\Zv),
\label{eq:wyner_learning_final}
\end{align}
where
\begin{align*}
\Dc_{\model}^{\xyzuv}
&\defeq\Dsym(\qvar(\xv,\yv,\zv,\uv,\vv), \pmodel(\xv,\yv,\zv,\uv,\vv))
\numberthis
\label{eq:obj_matching_model_xyzuv}
\end{align*}
is the divergence between the variational distribution and the model distribution over $(\xv,\yv,\zv,\uv,\vv)$ and $\lambda_{\model}^{\mathsf{CI}}> 0$ is a hyperparameter that trades off matching distributions and seeking succinct representation.

We first relax the equality constraint $\Dc_{\model}^{\mathsf{xyz}}=0$ with an inequality constraint $\Dc_{\model}^{\mathsf{xyz}}\le \eps$ for some $\eps>0$ and as in \cite{Zhao--Song--Ermon2018} to convert the problem~\eqref{eq:wyner_each_model_2} into an unconstrained Lagrangian form
\begin{align}\minimize ~\Dc_{\model}^{\mathsf{xyz}}+\lambda_{\model}^{\mathsf{CI}} I_{\model}(\Xv,\Yv;\Zv),
\label{eq:wyner_learning_xyz}
\end{align}
where the reciprocal of a Lagrange multiplier $\lambda_{\model}^{\mathsf{CI}}>0$ controls the model CI of $\pmodel(\xv,\yv,\zv)$.
We then introduce an additional variational relaxation step to train the style extractors $q_\phi(\uv|\zv,\xv)$ and $q_\phi(\vv|\zv,\yv)$.
That is, by the monotonicity of $f$-divergences (see Proposition~\ref{prop:fdiv_monotone} in Appendix), we have 
\begin{align}
\Dc_{\model}^{\mathsf{xyz}}
&\defeq \Dsym(\qdata(\xv,\yv,\zv),\pmodel(\xv,\yv,\zv))\nonumber\\
&\le \Dsym(\qvar(\xv,\yv,\zv,\uv,\vv), \pmodel(\xv,\yv,\zv,\uv,\vv))\nonumber\\
&\eqdef \Dc_{\model}^{\xyzuv},
\label{eq:var_up_bound}
\end{align}
where the equality holds if and only if the composite variational encoders 
\[
\qvar(\zv,\uv,\vv|\xv,\yv)=q_\phi(\zv|\xv,\yv)q_\phi(\uv|\zv,\xv)q_\phi(\vv|\zv,\yv)
\]
match to the model posterior $\pmodel(\zv,\uv,\vv|\xv,\yv)$. 
After all, we obtain \eqref{eq:wyner_learning_final} 
as the final relaxed optimization problem for each model distribution $\pmodel\in\{\pjoint, \pxtoy, \pytox\}$.

\begin{remark}[On the variational common information]
An acute reader may suggest to simply control the CI regularization by a single term $I_{\var}(\Xv,\Yv;\Zv)$, the variational CI.
Indeed, given that $q_\phi(\zb|\xb,\yv)$ is succinctly learned based on \eqref{eq:wyner_recall}, one could expect that the rest of the model components would be encouraged to be consistent with the succinctly learned $q_\phi(\zv|\xv,\yv)$, via the distribution matching term $\Dc_{\model}^{\xyzuv}$~\eqref{eq:var_up_bound}.
We empirically found, however, that  $I_{\var}(\Xv,\Yv;\Zv)$ as a regularization term
is not sufficiently effective to control the model CI's and leads to unstable training,
compared to directly using the model CI $I_{\model}(\Xv,\Yv;\Zv)$ for regularization.
Intuitively, this phenomenon may be attributed to %
imperfect distribution matching, which is enforced by minimizing $\Dc_{\model}^{\xyzuv}$ in our framework,
due to a lack of samples, a limited expressivity of the parametric models, an imperfect training, or their combinations.
\end{remark}

\subsection{Auxiliary Objectives}
\label{sec:objectives_auxiliary}
Note that learning a succinct common representation becomes meaningful only when a good degree of consistency between the models and data can be assured.
In principle, solving the optimization problem~\eqref{eq:wyner_learning_final}
may suffice for training the target generative models with a succinct common representation. 
Since, however, such non-convex optimization problems are hard to solve in general, distribution matching may not take place to begin with. 
Hence, in this section, we additionally introduce a set of auxiliary objectives that can considerably improve the degree of distribution matching.

\subsubsection{Reconstruction Losses} 
With the variational encoders, we can further guide the training by imposing certain \emph{reconstruction consistency} in the model, so that the optimization of the encoders and decoders is over a restricted function space that conforms to the consistency.
Note that the model trained with the reconstruction loss terms below can be viewed as a form of \emph{autoencoders}.

For the joint model $\pjoint$, similar to the reconstruction in autoencoders, it is natural to desire that the decoders $\xv_\th(\zv,\uv), \yv_\th(\zv,\vv)$ map the inferred representations $(\zv_o,\uv_o,\vv_o)\sim\qvar(\zv,\uv,\vv|\xv_o,\yv_o)$ for a given pair $(\xv_o,\yv_o)$ back to $(\xv_o,\yv_o)$; hence, we aim to minimize the \emph{joint data reconstruction losses} defined as
\begin{align*}
\Rc_{\mathsf{xy\to x}}\defeq \E_{\qvar(\xv,\yv,\zv,\uv,\vv)p_\th(\hat{\xv}|\zv,\uv)}[d_{\mathsf{x}}(\xv,\hat{\xv})]
\numberthis
\label{eq:obj_recon_joint_x}
\end{align*}
and the symmetrically defined $\Rc_{\mathsf{xy\to y}}$ for some dissimilarity functions $d_{\mathsf{x}}(\xv,\hat{\xv})$ and $d_{\mathsf{y}}(\yv,\hat{\yv})$.

For the conditional model $\pxtoy$, we consider the following consistency on the data space. 
Given a data pair $(\xv_o,\yv_o)$, we first draw a common representation $\zv_o\sim q_\phi(\zv|\xv)$ only from $\xv_o$ and find a local representation of $\yv_o$ conditioned on $\zv_o$, \ie $\vv_o\sim q_\phi(\vv|\zv_o,\yv_o)$. Then, we expect the decoder $\yv_\th(\zv,\vv)$ to reconstruct $\yv_o$ from the representation $(\zv_o,\vv_o)$, which leads to the definition of the \emph{conditional reconstruction loss} 
\begin{align*}
\Rc_{\xtoy}\defeq
\E_{\qdata(\xv,\yv)q_\th(\zv|\xv)q_\phi(\vv|\zv,\yv)p_\th(\hat{\yv}|\zv,\vv)}[d_{\mathsf{y}}(\yv,\hat{\yv})]
\numberthis
\label{eq:obj_recon_cond_x2y}
\end{align*} 
from $\xv$ to $\yv$; the other direction $\Rc_{\ytox}$ is symmetric.

\subsubsection{When Learning Joint and Conditional Models Simultaneously: Common Latent Space Matching Losses}
When we wish to train a single model that can perform every direction of inference (\ie joint and both ways of conditional generation), it is important to enforce the induced aggregated posteriors of the model encoders $q_\th(\zv|\xv)$ and $q_\th(\zv|\yv)$ in the conditional models to be consistent with the prior distribution $p_\th(\zv)$, so that they can share the common latent space over $\zv$. 
That is, we wish to match the aggregated posterior $\pmodel(\zv)$ to the prior $p_\th(\zv)$ for $\pmodel\in\{\pxtoy, \pytox\}$.
Since it is only enforced indirectly by the distribution matching losses $\Dc_{\model}^{\xyzuv}$, we further introduce the \emph{latent matching objectives}
\begin{align*}
\Mc_{\model}&\defeq \Dsym(\pmodel(\zv),p_\th(\zv)).
\numberthis
\label{eq:obj_matching_z}
\end{align*}
We remark that this consistency is also enforced by minimizing the distribution matching objective $\Dc_{\model}^{\xyzuv}$.
More precisely, by the monotonicity of $f$-divergences (Proposition~\ref{prop:fdiv_monotone}), we have $\Mc_{\model}\le\Dc_{\model}^{\xyzuv}$. 
Hence, additionally introducing the term $\Mc_{\model}$ should be understood as further encouraging the consistency between the model aggregated posterior $p_\model(\zv)$ and the prior, so that it improves the quality of downstream tasks.
We empirically found this objective especially helpful when learning with a paired data from two different domains such as an image-caption dataset, where the hardness of learning the target modalities is imbalanced.

\subsubsection{When Learning Both Conditional Models Simultaneously: Cross Matching Loss and Marginal Reconstruction Losses}
In some applications such as the cross-domain retrieval task (Remark~\ref{rem:retrieval}), we are only interested in learning conditional models of both directions such that they share the same common latent space, without learning the joint distribution.
In this case, we find that matching the two conditional distributions, \ie minimizing 
\begin{align*}
\Dc_{\xtoytox}^{\xyzuv}
&\defeq \Dsym(\pxtoy(\xv,\yv,\zv,\uv,\vv),\pytox(\xv,\yv,\zv,\uv,\vv)),
\numberthis
\label{eq:obj_matching_cross_xyzuv}
\end{align*}
can improve the quality of representation.

Another auxiliary term we find helpful in this scenario is the \emph{marginal} reconstruction losses
\begin{align*}
\Rc_{\mathsf{y\to y}}
\defeq \E_{\qdata(\yv)q_\th(\zv|\yv)q_\phi(\vv|\zv,\yv)p_\th(\hat{\yv}|\zv,\vv)}[d_{\mathsf{y}}(\yv,\hat{\yv})]
\numberthis
\label{eq:obj_recon_marg_x}
\end{align*}
and the symmetrically defined $\Rc_{\mathsf{x\to x}}$, which naturally arise when matching two conditional models $\pxtoy$ and $\pytox$.

\subsection{The Final Objective}
All the losses introduced so far are summarized in Table~\ref{tab:objectives}.
In experiments, we optimized a weighted combination of those objectives, and tune the weights as hyperparameters based on the \emph{hardness} of learning the corresponding modalities.

\begin{table*}[tb]
    \centering
    \caption{Summary of the objectives for training the variational Wyner model. 
    The objectives in the square brackets $[\cdot]$ with tilde notation are the corresponding discriminator objectives; see Section~\ref{sec:training_var_dre}.
    For the sake of easy reference, we indicate the definition for each objective term.
    The shaded objectives are the main objectives introduced in Section~\ref{sec:objectives_main}, which are derived from the relaxed Wyner optimization problem~\eqref{eq:wyner_learning_final}. 
    The rest are auxiliary objectives defined in Section~\ref{sec:objectives_auxiliary}.
    }
    {
    \begin{tabular}{ccccc}
    \toprule 
    Type (key)
    & Distribution matching
    & CI regularization
    & Reconstruction
    & Latent matching
    \\
    \midrule
        Joint $(\to\xv\yv)$
        & \ccell[gray]{0.9}{$\Dc_{\joint}^{\xyzuv}~\eqref{eq:obj_matching_model_xyzuv}~[\tilde{\Dc}_{\joint}^{\xyzuv}~\eqref{eq:disc_obj_matching_toxy}]
        $} & \ccell[gray]{0.9}{$I_{\joint}~\eqref{eq:obj_ci_model}~[\tilde{I}_{\joint}~\eqref{eq:disc_obj_ci_model}]$} & $\Rc_{\mathsf{xy\to x}}$, $\Rc_{\mathsf{xy\to y}}$~\eqref{eq:obj_recon_joint_x}
        &
        -\\
        Cond. $(\xv\to\yv)$
        & \ccell[gray]{0.9}{$\Dc_{\xtoy}^{\xyzuv}~\eqref{eq:obj_matching_model_xyzuv}~[\tilde{\Dc}_{\xtoy}^{\xyzuv}~(\text{cf.~}\ref{eq:disc_obj_matching_toxy})]
        $} & \ccell[gray]{0.9}{$I_{\xtoy}~\eqref{eq:obj_ci_model}~[\tilde{I}_{\xtoy}~\eqref{eq:disc_obj_ci_model}]$} & $\Rc_{\xtoy}$~\eqref{eq:obj_recon_cond_x2y} &
        $\Mc_{\xtoy}~\eqref{eq:obj_matching_z}~[\tilde{\Mc}_{\xtoy}~\eqref{eq:disc_obj_matching_latent}]$\\
        Cond. $(\xv\to\yv)$
        & \ccell[gray]{0.9}{$\Dc_{\ytox}^{\xyzuv}~\eqref{eq:obj_matching_model_xyzuv}~[\tilde{\Dc}_{\ytox}^{\xyzuv}~(\text{cf.~}\ref{eq:disc_obj_matching_toxy})]
        $} & \ccell[gray]{0.9}{$I_{\ytox}~\eqref{eq:obj_ci_model}~[\tilde{I}_{\ytox}~\eqref{eq:disc_obj_ci_model}]$} & $\Rc_{\ytox}$~\eqref{eq:obj_recon_cond_x2y} &
        $\Mc_{\ytox}~\eqref{eq:obj_matching_z}~[\tilde{\Mc}_{\ytox}~\eqref{eq:disc_obj_matching_latent}]$\\
    \midrule
        Cond. $(\xv\leftrightarrow\yv)$
        & $\Dc_{\xtoytox}^{\xyzuv}~\eqref{eq:obj_matching_cross_xyzuv}~[\tilde{\Dc}_{\xtoytox}^{\xyzuv}~(\text{cf.~}\ref{eq:disc_obj_matching_toxy})]$ 
        & - 
        & $\Rc_{\mathsf{x\to x}},\Rc_{\mathsf{y\to y}}$~\eqref{eq:obj_recon_marg_x}
        & -\\
    \bottomrule
    \end{tabular}}
    \label{tab:objectives}
\end{table*}

\section{Training Method}
\label{sec:training}
As alluded to earlier, we assume implicit generative models with deterministic decoders, the densities of the model distributions are not computable.
Hence, minimizing the objective functions introduced above requires a GAN-like adversarial technique. 
In particular, we adopt a technique proposed by \citet{Pu--Wang--Henao--Chen--Gan--Li--Carin2017SymmetricVAE}, which we call the \emph{variational density ratio estimation}. 
After all, the proposed training scheme can be viewed as an adversarial learning method of the variational Wyner model.
We also illustrate some tricks that were empirically effective for training in our experiments.

\subsection{Training with Variational Density Ratio Estimation}
\label{sec:training_var_dre}
Note that all the objective terms proposed above are in the form of either $\Dsym(p(\sv),q(\sv))$ for distribution matching and $\Dkl{p(\sv)}{q(\sv)}$ for mutual information, except the reconstruction losses.
To training with these divergence terms, we approximate the divergence terms by estimating the density ratio $p(\sv)/q(\sv)$ via an adversarial technique based on a variational characterization of the Jensen--Shannon divergence; namely, we use the optimal solution $r(\sv)$ of the following maximization problem
\begin{align}
D_{\mathsf{JS}}(p(\sv),q(\sv))
=&\max_{r(\sv)}  \psi_{\mathsf{JS}}(r(\sv);p(\sv),q(\sv)),
\label{eq:jsd_variational}
\end{align}
where we define
\begin{align}
&\psi_{\mathsf{JS}}(r(\sv);p(\sv),q(\sv))\nonumber\\
&\defeq \E_{p(\sv)}[\log \sigma(\log r(\sv))]
+ \E_{q(\sv)}[\log \sigma(-\log r(\sv))],
\label{eq:jsd_var}
\end{align}
to estimate the density ratio $p(\sv)/q(\sv)$, since the maximum of \eqref{eq:jsd_variational} is attained if and only if $r^*(\sv)\equiv p(\sv)/q(\sv)$. 
Here, $\sigma(x)=1/(1+e^{-x})$ denotes the sigmoid function.
Note that this is equivalent to the discriminator objective of the original generative adversarial networks (GANs)~\citep{Goodfellow--Pouget-Abadie-Mirza--Xu--Warde-Farley--Ozair--Courville--Bengio2014}, where $D(\sv)\defeq \sigma(\log r(\sv))\in [0,1]$ is called the \emph{discriminator}; in this work we view $r(\sv)=\exp(\sigma^{-1}(D(\sv)))\in (0,\infty)$ as a density ratio estimator but also call a discriminator, slightly abusing the terminology. 
While, in principle, the variational characterization of any $f$-divergence by \citet{Nguyen--Wainwright--Jordan2010} may be used to train a density estimator in a similar spirit of $f$-GANs~\citep{Nowozin--Botond--Tomioka2016}, we empirically observed that other choices of $f$-divergences such as one-sided KL divergences and $\chi^2$-divergences result in unstable training (data not shown).

As in a standard GAN training procedure, we alternate between training the variational Wyner model components and training the discriminators batch-by-batch, freezing one while training the other. 
When training the variational Wyner model, we freeze the density ratio estimators and estimate $\Dsym(p(\sv),q(\sv))$ by plugging in the approximate ratio $r(\sv)$ assuming that $r(\sv)\approx p(\sv)/q(\sv)$, \ie
\[
\Dsym(p(\sv),q(\sv))
\approx \E_{p(\sv)}[\log r(\sv)]-\E_{q(\sv)}[\log r(\sv)].
\]

Hence, for each distribution matching objective $\Dc_{\mathsf{key}}^\xyzuv$ for $\mathsf{key}\in\{\joint,\xtoy,\ytox,\xtoytox\}$, we introduce a corresponding density ratio estimator $r_{\mathsf{key}}(\xv,\yv,\zv,\uv,\vv)$ and optimize it by the discriminator objective $\tilde{\Dc}_{\mathsf{key}}^\xyzuv$, where,
\eg
\begin{align*}
\tilde{\Dc}_{\joint}^\xyzuv
\defeq \psi_{\mathsf{JS}}(r_{\joint}; \qvar, \pjoint).
\numberthis
\label{eq:disc_obj_matching_toxy}
\end{align*}
For a latent matching objective $\Mc_{\model}$, we train a discriminator $r_{\model}^{\mathsf{latent}}(\zv)$ by maximizing
\begin{align*}
\tilde{\Mc}_{\model}
\defeq& \psi_{\mathsf{JS}}(r_{\model}^{\mathsf{latent}}(\zv); \pmodel(\zv), p_\th(\zv)),
\numberthis
\label{eq:disc_obj_matching_latent}
\end{align*}
for each $\model\in\{\xtoy, \ytox\}$.
The mutual information $I_{\model}(\Xv,\Yv;\Zv)$ term can be handled by the same technique, \ie
training a discriminator $r_{\model}^{\mathsf{CI}}(\xv,\yv,\zv)$ by maximizing 
\begin{align*}
\tilde{I}_{\model}
&\defeq 
\psi_{\mathsf{JS}}(r_{\model}^{\mathsf{CI}}; \pmodel(\xv,\yv,\zv), \pmodel(\xv,\yv)\pmodel(\zv)),
\numberthis
\label{eq:disc_obj_ci_model}
\end{align*}
so that $r_{\model}^{\mathsf{CI}}(\xv,\yv,\zv)\approx \pmodel(\xv,\yv,\zv)) / \pmodel(\xv,\yv)\pmodel(\zv)$, and approximate the mutual information by the same plug-in approach:
\begin{align*}
I_{\model}(\Xv,\Yv;\Zv)
&=\E_{\pmodel(\xv,\yv,\zv)}\Bigl[\log \frac{\pmodel(\xv,\yv,\zv)}{\pmodel(\xv,\yv)\pmodel(\zv)}\Bigr]\\
&\approx \E_{\pmodel(\xv,\yv,\zv)}[\log r_{\model}^{\mathsf{CI}}(\xv,\yv,\zv)].\end{align*}
In the minibatch training of density ratio estimators for CI estimation, in order to to sample from a product distribution $\pmodel(\xv,\yv)\pmodel(\zv)$, we first draw $(\xv,\yv,\zv)\sim\pmodel(\xv,\yv,\zv)$ from the joint distribution and then simply permute $\zv$ over the batch dimension as a proxy to independent sampling.

\subsection{The Final Discriminator Objective}
In experiments, to train the discriminators, we simply added the corresponding objective terms without additional weights. 

\subsection{Additional Tricks for Training}
To make the training scheme more computationally efficient and stable, we made several important design choices, 
including (1) a shared joint data feature map among discriminators, (2) deterministic parameterization of encoders, and (3) the instance noise trick~\citep{Sonderby--Caballero--Theis--Shi--Huszar2016}. 

\subsubsection{Shared Feature Map in Discriminators}
In principle, for each pair of distributions whose density ratio is required to be estimated, we need a density ratio estimator $r_{\model}(\xv,\yv,\zv,\uv,\vv)$ for distribution matching or $r_{\model}^{\mathsf{CI}}(\xv,\yv,\zv)$ for CI regularization or $r_{\model}^{\mathsf{latent}}(\zv)$ for latent matching. 
To reduce the size of the discriminator network, in our implementation we use a single joint feature map $f(\xv,\yv)$ which maps the pair $(\xv,\yv)$ to a feature vector, and every density ratio estimator that takes $(\xv,\yv)$ as an argument is of the form either $r_{\model}(f(\xv,\yv),\zv,\uv,\vv)$ or $r_{\model}^{\mathsf{CI}}(f(\xv,\yv),\zv)$.

\subsubsection{Deterministic Encoders}
Following the standard practice, we approximate the proposed objectives in Table~\ref{tab:objectives} which are expectations over model distributions by a Monte Carlo approximation and plug-in it to a gradient-based optimization algorithm.
Note, however, that sampling distributions and taking gradients with respect to a parameter of the sampling distribution may cause \emph{biased} gradient estimates, since, in general,
\[
\nabla_\th \E_{p_\th(\sv)}[f_\th(\sv)]\neq  \E_{p_\th(\sv)}[\nabla_\th f_\th(\sv)].
\]
A possible detour is to deploy diagonal Gaussian encoders used in VAEs to invoke the reparametrization trick~\citep{Kingma--Welling2014}.
In this work, even simpler, we parameterize all the encoders (\ie variational encoders $q_\phi(\zv|\xv,\yv)$, $q_\phi(\uv|\zv,\xv)$, $q_\phi(\vv|\zv,\yv)$ and model encoders $q_\th(\zv|\xv)$, $q_\th(\zv|\yv)$) by \emph{deterministic} mappings, which can be viewed as the limiting version of the reparameterization trick with vanishing variances.

\subsubsection{Instance Noise Trick}
Since our encoders and decoders are all deterministic, they define \emph{degenerate} model distributions, on which the divergences and mutual information terms may not be properly defined due to disjoint support of paired distributions.
This is a well-known issue of implicit generative models, and we adopt the well-known \emph{instance noise trick} of \citet{Sonderby--Caballero--Theis--Shi--Huszar2016} from the GAN literature.
Namely, we add small Gaussian noise to all inputs of the discriminators (density estimators), which is equivalent to replacing any distribution $p(\xv,\yv,\zv,\uv,\vv)$ of our consideration with that convolved with a Gaussian kernel. 
We empirically observed that this trick is effective, but a proper tuning of the level of Gaussian noise is crucial in stabilizing the training procedure while not blurring out the distributions of our interest.

\section{Related Work}
\label{sec:related}

\subsection{On Wyner's CI and Related Measures}
Wyner's CI was first studied by \citet{Wyner1975} to investigate the problem of distributed simulation of two discrete random sources and distributed compression in the so-called Gray--Wyner network~\citep{Gray--Wyner1974}. 
\citet{Witsenhausen1976} established a lower bound on this quantity and studied its computability.
\citet{Cuff2013} established the role of Wyner's CI in its conditional counterpart, \ie the channel synthesis problem.
Later, \citet{Xu--Liu--Chen2016} studied the quantity for a pair of continuous random variables, 
and provided its operational justification in the distributed \emph{lossy} compression setting.

Recently, Wyner's common information has received a lot of attention in the information theory literature, especially in the context of its application for extracting correlation between dependent variables. 
A recent line of theoretical work includes a local characterization of Wyner's CI~\citep{Huang--Xu--Zheng--Wornell2020} and an alternative, Wyner's-CI-based procedure for canonical correlation analysis~\citep{Sula--Gastpar2021}.

There exist several other related dependence measures for a pair of random variables $(X,Y)$ in information theory.
The mutual information $I(X;Y)$ has significant roles and concrete operational meanings in information theory and statistics, including source and channel coding problems and hypothesis testing problems; see, \eg \citep{Cover--Thomas2006}.
We remark, however, that Wyner's common information $J(X;Y)$ is in general different from the more famous quantity of mutual information $I(X;Y)$. 
It is easy to prove that $0\le I(X;Y) \le J(X;Y)$. In general, the inequalities can be strict, but when $X$ and $Y$ are independent, $I(X;Y)=J(X;Y)=0$.

The G\'ac--K\"orner--Witsenhausen common information $K(X;Y)$~\citep{Gacs--Korner1973} is defined to be the maximum number of common bits per symbol that can be independently extracted from $X$ and $Y$. 
While it has several applications in secret key generation, it is known that the notion is rather restrictive in the sense that $K(X;Y)$ becomes positive only for limited cases~\citep{Gacs--Korner1973,Witsenhausen1975}. 
Moreover, this quantity can be defined only for discrete random variables.

The Hirschfeld--Gebelein--R\'enyi (HGR) maximal correlation~\citep{Hirschfeld1935,Gebelein1941,Renyi1959} is a nonlinear generalization of Pearson correlation coefficient. The HGR maximal correlation is originally defined for a pair of scalar random variables, but it was generalized to quantify a measure of dependence between high-dimensional random vectors; see \citep{Michaeli--Wang--Livescu2016}.
There also exists a line of work on the maximal correlation and its applications in machine learning from an information-theoretic view;
see a recent paper by \citet{Huang--Makur--Wornell--Zheng2019} for an overview on the recent theoretical breakthroughs.

For a more broad treatment on this subject, we refer an interested reader to a recent monograph by \citet{Yu--Tan2022}.

\subsection{Existing Information-Theoretic Approaches}
In this section, we provide an in-depth discussion on two information theoretic approaches~\citep{Tishby--Pereira--Bialek1999,Hwang--Kim--Hong--Kim2020IIAE}, elaborating the philosophical differences compared to our approach.

\subsubsection{Information Bottleneck Principle}
The information bottleneck (IB) principle (or method)~\citep{Tishby--Pereira--Bialek1999} is a widely known information theoretic approach in representation learning.
This approach is usually considered for \emph{discriminative tasks}, i.e., when the target variable $\Yv$ is a function of $\Xv$ and/or even discrete. 
Motivated by lossy compression, the IB principle proposes to find a \emph{compressed} representation $\Zv$ from the input variable $\Xv$ (i.e., $q_\th(\zv|\xv)$) while maximizing the relevance of $\Zv$ in \emph{predicting} the target variable $\Yv$ as the minimizer of the optimization problem
\[\minimize_{q_\th(\zv|\xv)} I(\Xv;\Zv)- \b I(\Yv;\Zv),
\]
where $(\Xv,\Yv,\Zv)\sim \qdata(\xv,\yv)q_\th(\zv|\xv)$ and $\b>0$.

Indeed, the IB principle and the proposed framework are suitable to discriminative tasks and generative tasks, respectively, while they fail to define a good representation structure in the other respective cases.
First, consider a discriminative task such as classification, where typically there is a near functional relationship $\Yv\approx f(\Xv)$ between $\Xv$ and $\Yv$. 
The proposed framework principle does not posit an interesting structure in this case, since the trivial choice $\Zv=\Yv$ makes $\Xv$ and $\Yv$ independent and achieves the minimum $I(\Xv,\Yv;\Zv)=H(\Yv)$, whereas the IB principle defines a series of representations of different levels of compression controlled by $\b$.
Secondly, for a generative task where the pair $(\Xv,\Yv)$ has many-to-many relationship, guessing $\Yv$ based on $\Zv$ as a representation of $\Xv$, the symmetric Markov assumption $\Xv\leftrightarrow\Zv\leftrightarrow\Yv$ of our approach is more appropriate than $\Zv\leftrightarrow\Xv\leftrightarrow\Yv$ of IB; crucially, under the Markov chain $\Zv\leftrightarrow\Xv\leftrightarrow\Yv$, $\Yv$ is not conditionally independent of $\Xv$ given $\Zv$ in general.
We summarize the differences in Table~\ref{table:IB}. 

In Appendix~\ref{app:minimal}, we discuss a connection from the notion of minimal sufficient statistics~\citep{Lehmann--Scheffe1950} to Wyner's optimization problem and the IB principle.

We finally remark that \citet{Alemi--Fischer--Dillon--Murphy2017} proposed to train a neural network classifier with a variational relaxation of the IB objective to seek a robust representation and a few variations of this work were proposed to find an invariant factors of a target $\Xv$ given an attribute $\Yv$~\citep{Gao--Brekelmans--Ver-Steeg--Galstyan2019,Song--Kalluri--Grover--Zhao--Ermon2019}.

\begin{table*}[t]
\centering
\caption{The variational Wyner model vs. the IB principle~\citep{Tishby--Pereira--Bialek1999}.}
\label{table:IB}
{\small
\begin{tabular}{rcc}
\toprule
& {\bf The variational Wyner model} & {\bf The IB principle} \\
\midrule
Motivating problem & \makecell{channel synthesis,\\distributed simulation} & \makecell{lossy compression,\\minimal sufficient statistics} \\
Probabilistic model & $\Xv\leftrightarrow\Zv\leftrightarrow\Yv$ & $\Zv\leftrightarrow\Xv\leftrightarrow\Yv$\\
Direction of inference & bidirectional & unidirectional \\
Measure of succinctness & $I(\Xv,\Yv;\Zv)$ & $I(\Xv;\Zv)$ \\
Measure of fit/relevance & $D(p,q)$ & $I(\Yv;\Zv)$ \\
Optimal quantity & $J(\Xv;\Yv)$ & N/A\\
\bottomrule
\end{tabular}}
\end{table*}

\subsubsection{Interactive Information Maximization}
Recently, \citet{Hwang--Kim--Hong--Kim2020IIAE} proposed a new information-theoretic regularization principle to tackle the cross-domain disentanglement problem.
To seek a disentangled representation $(\Zv,\Uv,\Vv)$ under the variational distribution
\[
\qvar'(\xv,\yv,\zv,\uv,\vv)\defeq \qdata(\xv,\yv)q_\phi(\zv|\xv,\yv)q_\phi(\uv|\xv)q_\phi(\vv|\yv),
\]
they propose to maximize the \emph{interactive information among $\Xv$, $\Yv$, $\Zv$} to enforce $\Zv$ to capture a commonality of $(\Xv,\Yv)$, while minimizing $I(\Zv;\Uv)$ and $I(\Zv;\Vv)$ to enforce the representations $(\Zv,\Uv,\Vv)$ to be independent. Here, the interactive information among $\Xv$, $\Yv$, $\Zv$ is defined as
\[I(\Xv;\Yv;\Zv)\defeq I(\Xv;\Zv)-I(\Xv;\Zv|\Yv)\] 
and it is symmetric in $(\Xv,\Yv,\Zv)$~\citep{McGill1954}.
After all, they proposed to minimize a variational upper bound of a weighted combination of a distribution matching term
\[
\Dkl{\qvar'(\xv,\yv,\zv,\uv,\vv)}{\pjoint(\xv,\yv,\zv,\uv,\vv)}
\]
and $I(\Zv;\Uv)+I(\Zv;\Vv)-2I(\Xv;\Yv;\Zv)$.
We remark that since $I(\Zv;\Uv)$ and $I(\Zv;\Vv)$ terms should become 0 when the paired distributions exactly match, the essential driving force of cross-domain disentanglement in this framework is from the \emph{maximization} of the interactive information. 
In a stark contrast, we propose to \emph{minimize} the information captured in $\Zv$.
Further, this interactive-information maximization framework does not define an optimality criterion for a good representation as alluded to earlier, and the choice of weights in the objective term $I(\Zv;\Uv)+I(\Zv;\Vv)-2I(\Xv;\Yv;\Zv)$ is rather ad-hoc, based on a computational aspect of the final objective.

\subsection{Other Cross-Domain Disentanglement Models and Bimodal Generative Models}
One of the most closely related work in the deep learning literature is the cross-domain disentanglement networks proposed by \citet{Gonzalez-Garcia--VanDeWeijer--Bengio2018}.
Similar to the variational Wyner model, their model also aims to decompose a joint representation into \emph{shared} (common) and \emph{exclusive} (local) representations explicitly. 
The crucial difference is that, in \citep{Gonzalez-Garcia--VanDeWeijer--Bengio2018}, one of the key components for disentanglement is the use of a \emph{gradient reversal layer}~\citep{Ganin--Lempitsky2015},
while the variational Wyner model forces to learn succinct information using the CI regularization terms, towards learning the optimally succinct representation characterized by \eqref{eq:wyner}.

The Wyner model can be viewed as a generalization of the probabilistic model (i.e., encoder and decoder) assumed in two existing joint variational autoencoders (VAEs)--- JVAE~\citep{Vedantam--Fischer--Huang--Murphy2018TELBO} and JMVAE~\citep{Suzuki--Nakayama--Matsuo2016JMVAE}---as alluded in the previous paragraph. 
These models implement a similar idea of performing joint and conditional generation tasks via a symmetric Markov chain $\Xv\leftrightarrow\Wb\leftrightarrow\Yv$, where $\Wb$ is the \emph{joint representation} of $(\Xv,\Yv)$. 
In other words, these models can be derived by removing the local variables $\Uv$ and $\Vv$ in the variational Wyner model.

The same decoder structure of the variational Wyner model with the ``shared'' ($\Zv$) and the ``private'' ($\Uv,\Vv$) latent variables has been also studied in the context of multi-view learning~\citep{Shon--Grochow--Hertzmann--Rao2006,Ek--Rihan--Torr--Rogez--Lawrence2008,Salzmann--Ek--Urtasun--Darrell2010,Damianou--Ek--Titsias--Lawrence2012} mostly based on a linear analysis such as canonical correlation analysis (CCA).
More recently, variational CCA-private (VCCA-private)~\citep{Wang--Yan--Lee--Livescu2016DeepVCCA} was proposed to learn the decoder model with variational encoders $q_\th(\zv|\xv)$, $q_\phi(\uv|\xv)$, and $q_\phi(\vv|\yv)$, with the encoder model $q_\phi(\zv,\uv,\vv|\xv,\yv)=q_\th(\zv|\xv)q_\phi(\uv|\xv)q_\phi(\vv|\yv)$ to directly capture the conditional model from $\Xv$ to $\Zv$ to $\Yv$.
On the other hand, the variational Wyner model relies on the conditional independence structure $q_\phi(\zv,\uv,\vv|\xv,\yv)=q_\phi(\zv|\xv,\yv)q_\phi(\uv|\zv,\xv)q_\phi(\vv|\zv,\yv)$, which is naturally induced by the decoder model; see Remark~\ref{rem:conditional_independence}.
We argue that this choice of encoder model in our variational Wyner model may capture better semantic meaning of the local (private) random variables $\Uv$ and $\Vv$, thereby leading a better generative performance; see, e.g., Fig.~\ref{fig:mnist_svhn}.

Conditional VAE (CVAE)~\citep{Sohn--Lee--Yan2015CVAE} directly models the conditional distribution $\qdata(\yv|\xv)$, obtained by simply conditioning every component in the vanilla VAE for $\qdata(\yv)$ with the conditioning variable $\Xv$. 
If $\Yv$ is an image and $\Xv$ is an attribute of the image, a latent representation $\Vv$ in CVAE needs to capture the redundant information of $\Yv$, which is not contained in $\Xv$, i.e., style information of $\Yv$ given $\Xv$.
The variational Wyner model can be viewed as a combination of two CVAEs with $\Zv$ as a common conditioning variable, being capable of bidirectional sampling in its nature.
Yet, if $\Xv$ is high-dimensional, the conditional models like CVAE in general tend to overfit the input data of $\Xv$~\citep{Dutordoir--Salimbeni--Hensman--Deisenroth2018}. 
To address this problem, a subsequent related work, bottleneck conditional density estimation (BCDE)~\citep{Shu--Bui--Ghavamzadeh2017BCDE}, proposed to learn joint and conditional VAE models simultaneously by softly tying the parameters of the two models for regularization.
We note that the variational Wyner model naturally addresses such problem by using a unified single probabilistic model for both joint and conditional distribution learning, finding a succinct common representation $\Zv$ for regularization.

\section{Experiments}\label{sec:exp}

We empirically demonstrate the power of the proposed approach with synthetic and real-world datasets.
We parameterized all model components by deep neural networks.
Details of the experiments such as training schemes, network architectures, and evaluation metrics are deferred to Appendix~\ref{app:exp_settings}.
We performed most of the experiments over the Triton Shared Computing Cluster~\citep{TSCC2022}.
The code is available online~\footnote{\url{https://github.com/jongharyu/wyner-model}}.

\subsection{MNIST--SVHN Add-One Dataset}
To show the effect of information decomposition in our model, we first considered a synthetic image-image pair dataset constructed from MNIST~\citep{LeCun1998mnist} and SVHN~\citep{Netzer--Wang--Coates--Bissacco--Wu--Ng2011} datasets, similar to \citet{Shi--Siddharth--Paige--Torr2019MMVAE}. Here, we randomly picked an MNIST image $\Xv_i$ of label $\ell_i\in \{0,\ldots,9\}$ and paired with four randomly picked SVHN images of label  $(\ell_i+1) \mod 10$; we call the resulting dataset the MNIST--SVHN add-one dataset. Note that the images are paired only through their labels, and clearly the common information structure we seek is the underlying label of a pair.

We trained all the joint and conditional models simultaneously, with the objectives
\begin{align*}
&\Dc_{\joint}^{\xyzuv}+\Dc_{\xtoy}^{\xyzuv}+\Dc_{\ytox}^{\xyzuv}\\
&+\lambda^{\mathsf{CI}}(I_{\joint}+I_{\xtoy}+I_{\ytox}) \\
&+ \Rc_{\mathsf{xy\to x}} + \Rc_{\mathsf{xy\to y}} + \Rc_{\xtoy} + \Rc_{\ytox}
\end{align*} 
for training the variational Wyner model and
\begin{align*}
\tilde{\Dc}_{\joint}^{\xyzuv}+\tilde{\Dc}_{\xtoy}^{\xyzuv}+\tilde{\Dc}_{\ytox}^{\xyzuv}+\tilde{I}_{\joint}+\tilde{I}_{\xtoy}+\tilde{I}_{\ytox}
\end{align*} for training the discriminator.
We tried four different CI regularization weight $\lambda^{\mathsf{CI}}\in\{0,0.1,0.2,0.5,1\}$ to demonstrate the effect of the regularization for 25 epochs and the averaged $\ell_1$-distance over dimensions was used for the reconstruction loss functions.
The dimension of the latent space $(\Zv,\Uv,\Vv)$ was $(16, 16, 16)$.

\begin{figure*}[!tbh]
\centering
\includegraphics[width=\textwidth]{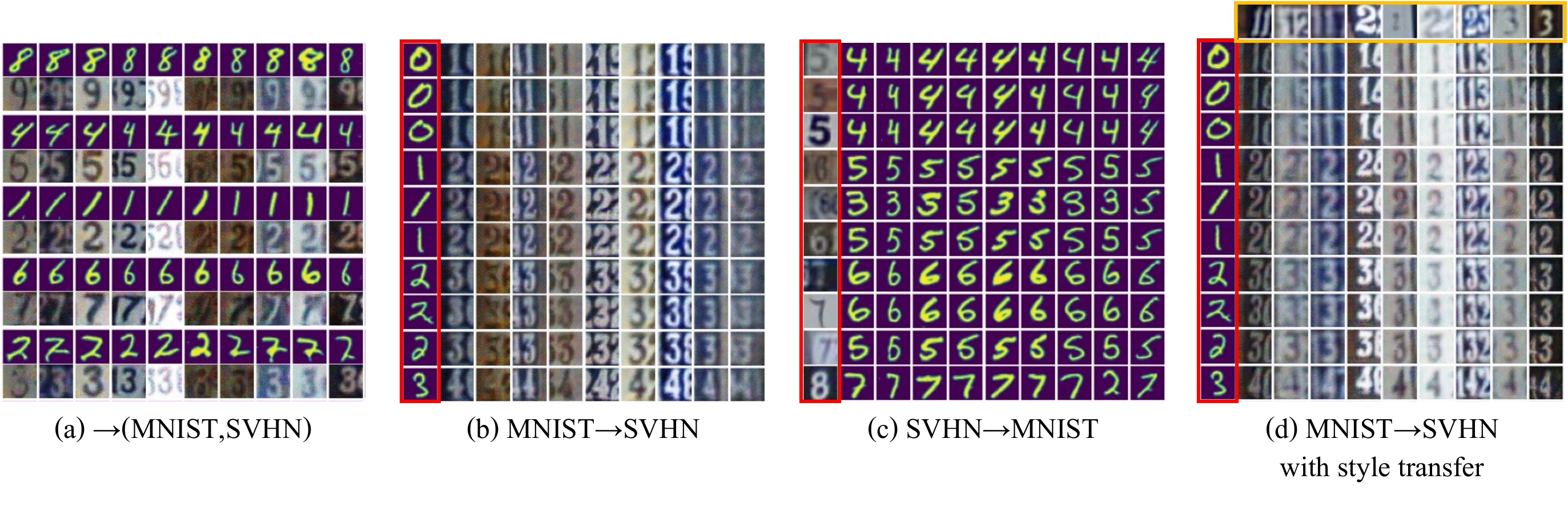}
\vspace{-3.em}
\caption{MNIST--SVHN add-one samples from the variational Wyner model. 
In (b)-(d), the images in the red boxes are inputs to the conditional models. 
In (d), the yellow box highlights the style reference images.}
\label{fig:mnist_svhn}
\end{figure*}

In Fig.~\ref{fig:mnist_svhn}, we present a few joint and conditional samples generated from the trained model with $\lambda^{\mathsf{CI}}=0.5$ at the end of training. In the figure, each row shares the same $\zv$, and each column shares the same $\uv$ and/or $\vv$. In particular, the top row of the last panel (d) shows the reference samples whose style are transferred downward along each column.
The samples clearly indicate that the learned model successfully disentangles the common and local representations. For example, in Fig.~\ref{fig:mnist_svhn}(b), in the first three rows, regardless of the specifics of the input MNIST images independent to their label 0, the generated samples coherently present the correct label 1 as well as sharing the same style fixed along each column.
Fig.~\ref{fig:mnist_svhn}(d) illustrates that using the local variational encoder $q_\phi(\uv|\zv,\xv)$, we can generate conditional samples given a fixed style extracted from a reference image; recall Remark~\ref{rem:style}.%

\begin{figure}%
\centering
\includegraphics[width=0.5\textwidth]{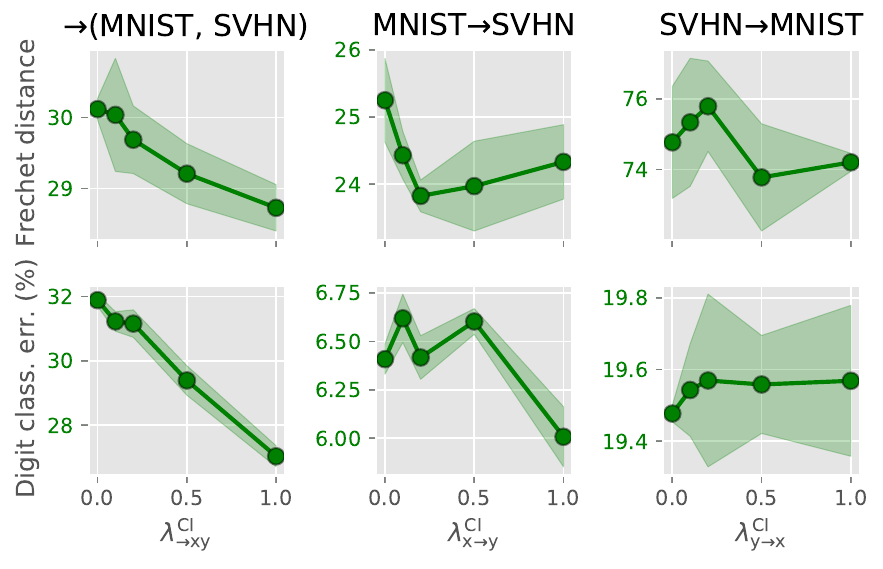}
\vspace{-1.5em}
\caption{A summary of numerical evaluations for MNIST--SVHN add-one dataset.
We ran five experiments with different random seeds and report the average scores. The shaded areas indicate the standard deviations.
}
\label{fig:mnist_svhn_numerical}%
\end{figure}
In order to numerically examine the effect of CI regularization in the model, we computed two metrics (1) custom Frechet distance (FD) scores and (2) classification accuracy of generated samples. 
For the Frechet distance scores, unlike the original proposal by \cite{Heusel--Ramsauer--Unterthiner--Nessler--Hochreiter2017TTUR} using the Inception network, we separately trained two fully convolutional autoencoders with MNIST and SVHN datasets, respectively, and used the bottleneck features to compute the first and second order statistics for computing the Frechet distance.
The digit classification errors were computed by pretrained classifiers for MNIST and SVHN. 
We refer the reader to the details for computing these metrics to Appendix.
While achieving lower values under both metrics are ideal, note that there exists a natural trade-off between them. 
For example, a joint generative model that only generates good images of the digit pairs (1,2) could achieve zero error in the classification error, but may suffer a large FD score.
On the other extreme, as a model tends to generate more diverse samples with different styles, it may be more prone to suffer a large error in the digit consistency.

The results are summarized in Fig.~\ref{fig:mnist_svhn_numerical}. 
As shown in the figure, in general, increasing $\lambda^{\mathsf{CI}}$ improves the quality of generated samples in terms of the smaller FD scores and improved the (estimated) digit accuracy. Note that the effect of the CI regularization is clear in $\to$(MNIST, SVHN) and MNIST$\to$SVHN, while it is
not clear in SVHN$\to$MNIST.
As an explanation to this phenomenon in SVHN$\to$MNIST, we remark that the learned model even with $\lambda_{\ytox}^{\mathsf{ci}}=0$ achieved a decent degree of disentanglement, which can be justified indirectly via the generated samples as shown in Fig.~\ref{fig:mnist_svhn}(c).
In general, for conditional models, we observed in our experiments that the CI regularization becomes more effective for the direction from a simpler modality (\eg MNIST) to a more complex one (\eg SVHN).

\subsection{CUB Image-Caption Dataset}
We further demonstrate that the proposed model can even learn a complex real-world image-caption dataset, following the same setting of \citep{Shi--Siddharth--Paige--Torr2019MMVAE}.
We used the Caltech-UCSD Birds (CUB) dataset~\citep{Wah--Branson--Welinder--Perona--Belongie2011CUB} that consists of 11,788 photos of birds, each of which is paired with 10 captions. To simplify the learning task, we translate the images into 2048-dimensional ResNet-101 features~\citep{He--Zhang--Ren--Sun2016}; to reconstruct a real image from feature, we retrieved the nearest neighbor in the feature space with respect to the Euclidean distance.

For this dataset, we found that learning the image modality $(\xv)$ from captions $(\yv)$ is harder than the other way around, and thus puts larger weights on learning the image modality. In particular, we used the following objective function with imbalanced weights
\begin{align*}
&\Dc_{\joint}^{\xyzuv}+\Dc_{\xtoy}^{\xyzuv}+\Dc_{\ytox}^{\xyzuv}\\
&+0.5(I_{\joint}+I_{\ytox}) \\
&+ 256(\Rc_{\mathsf{xy\to x}}+\Rc_{\ytox}+\Mc_{\ytox}) \\
&+ 8(\Rc_{\mathsf{xy\to y}} + \Rc_{\xtoy} +\Mc_{\xtoy})
\end{align*}
for training the variational Wyner model and \begin{align*}
\tilde{\Dc}_{\joint}^{\xyzuv}+\tilde{\Dc}_{\xtoy}^{\xyzuv}+\tilde{\Dc}_{\ytox}^{\xyzuv}+\tilde{I}_{\joint}+\tilde{I}_{\ytox}
\end{align*}
for training the discriminator. 
Note that we did not use the CI regularization term $I_{\xtoy}$ (from image to caption).
We used the $\ell_2^2$-distance for $\xv$ (image-feature space) and $(\zv,\uv,\vv)$ (latent space), and the categorical cross-entropy loss for $\yv$ (sentence space), all averaged over dimensions, for the reconstruction loss functions.
The dimension of the latent space $(\Zv,\Uv,\Vv)$ was $(16, 8, 8)$.

As a numerical evaluation, we computed the correlation scores of jointly and conditionally generated samples with respect to the training samples, via a canonical correlation analysis, following \citep{Massiceti--Dokania--Siddharth--Torr2018,Shi--Siddharth--Paige--Torr2019MMVAE}; we refer an interested reader to Appendix for details. After 20 epochs of training, we attained correlation scores \textbf{(0.303, 0.327, 0.318)} for joint, conditional generation ($\xv\to\yv$ and $\yv\to\xv$), respectively. 
Note that the correlation scores computed with the test dataset is 0.273 and the reported scores attained by \citet{Shi--Siddharth--Paige--Torr2019MMVAE} was (0.263, 0.104, 0.135).
We further report that without the CI regularization terms $0.5(I_{\joint}+I_{\ytox})$ the model fails to disentangle the representations (data not shown). 

\begin{figure*}
    \centering
    \includegraphics[width=\textwidth]{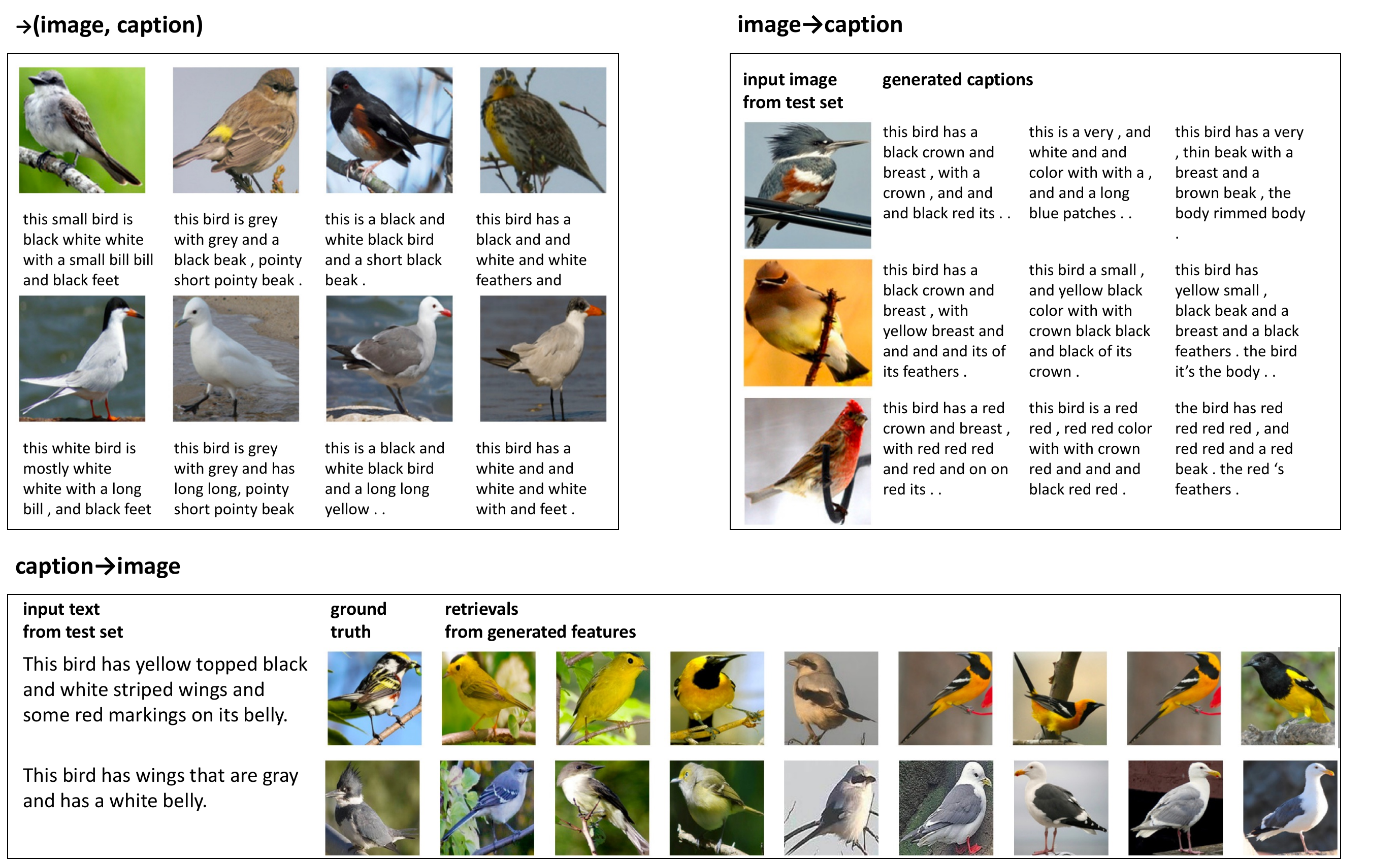}
    \vspace{-1em}
    \caption{Samples from the variational Wyner model trained with CUB Image-Caption dataset. Note that we generated image features and the shown images are retrieved based on the nearest features from the test data. 
    }
    \label{fig:cub_samples}
\end{figure*}

We present some examples of generated samples in Fig.~\ref{fig:cub_samples}, which show that the trained variational Wyner model indeed generates a variety of samples of high coherence. 

\subsection{Zero-Shot Sketch Based Image Retrieval}\label{sec:zs_sbir}
To demonstrate the utility of learned representations beyond generative modeling, 
we consider the \emph{zero-shot sketch based image retrieval} (ZS-SBIR) task proposed by \citet{Yelamarthi--Reddy--Mishra--Mittal2018}, where the goal is to construct a good retrieval model that retrieves relevant photos from a sketch, with a training set of no overlapping classes with a test set.

For this experiment, we borrowed the the same setting from \citet{Hwang--Kim--Hong--Kim2020IIAE}. 
We trained and evaluated our model with the Sketchy Extended dataset~\citep{Sketchy2016,Liu--Shen--Shen--Liu--Shao2017SketchyExtended}, which consists of total 75,479 sketches ($\Xv$) and 73,002 photos ($\Yv$) from 125 different classes. 
During training, we constructed a random pair of a photo and a sketch from a same class. For training and evaluation, a pretrained VGG16 network~\citep{Simonyan--Zisserman2014VGG} was used to extract features of the images. 
We used the pretrained VGG network and train-test splits for evaluation from the codebase\footnote{\url{https://github.com/AnjanDutta/sem-pcyc}} of \cite{Dutta--Akata2019SEM-PCYC}.
After training, we performed the retrieval task by the procedure illustrated in Remark~\ref{rem:retrieval}.

We trained our model only with conditional model components, as we only need to learn good model encoders $q_\th(\zv|\xv)$ and $q_\th(\zv|\yv)$. 
Specifically, we used the objectives
\begin{align*}
&\Dc_{\xtoy}^{\xyzuv}
+\Dc_{\ytox}^{\xyzuv}
+\Dc_{\xtoytox}^{\xyzuv}\\
&+\lambda^{\mathsf{CI}}(I_{\xtoy}+I_{\ytox}) \\
&+\lambda^{\mathsf{rec}}(\Rc_{\mathsf{\xtoy}} + \Rc_{\mathsf{\ytox}} + \Rc_{\mathsf{x\to x}} + \Rc_{\mathsf{y\to y}})
\end{align*}
for training the variational Wyner model and
\begin{align*}
&\tilde{\Dc}_{\xtoy}^{\xyzuv}+\tilde{\Dc}_{\ytox}^{\xyzuv}
+\tilde{I}_{\xtoy}+\tilde{I}_{\ytox}
\end{align*}
for training the discriminator. 
We used the $\ell_2^2$-distance averaged over dimensions for the reconstruction loss functions.
The dimension of the latent space $(\Zv,\Uv,\Vv)$ was $(64, 64, 64)$.

As a quantitative evaluation, we computed the Precision@100 (P@100) and mean average precision (mAP) scores on the test split; see Table~\ref{tab:sketchy}.

\begin{table}[tbh]
    \centering
    \caption{Evaluation of the ZS-SBIR task with the Sketchy Extended dataset.}%
    \begin{tabular}{c c c}
    \toprule
        Models & P@100 & mAP \\
     \midrule
        LCALE~\cite{Lin--Xu--Gao--Wang--Shen2020} & 0.583 & 0.476 \\
        IIAE~\cite{Hwang--Kim--Hong--Kim2020IIAE} & 0.659 & 0.573 \\
    \midrule
        Variational Wyner & \textbf{0.703} & \textbf{0.629} \\
    \bottomrule
    \end{tabular}
    \label{tab:sketchy}
\end{table}

The reported scores for the adversarially learned Wyner model was obtained with $\lambda^{\mathsf{CI}}=0.1$ and $\lambda^{\mathsf{rec}}=8$. We outperform the scores reported by \citet{Hwang--Kim--Hong--Kim2020IIAE}, who already demonstrated that their scores significantly improved upon the existing work tailored to extra information; for example, LCALE~\cite{Lin--Xu--Gao--Wang--Shen2020} incorporated word embedding during training.
The improvement corroborates the power of our approach in learning disentangled representations.\footnote{To make a fair comparison as possible, the latent dimension and network architecture of the variational Wyner model part were also chosen almost identical to the one used in \citep{Hwang--Kim--Hong--Kim2020IIAE}.}
For an ablation study, we trained our model with degenerate local encoders $q_\phi(\uv|\xv)$ and $q_\phi(\uv|\yv)$, \ie without conditioning with $\zv$, and achieved suboptimal scores (0.670,0.591); it justifies the design of our local encoders $q_\phi(\uv|\zv,\xv)$ and $q_\phi(\vv|\zv,\yv)$.

\begin{figure*}[ht]
    \centering
    \includegraphics[width=\textwidth]{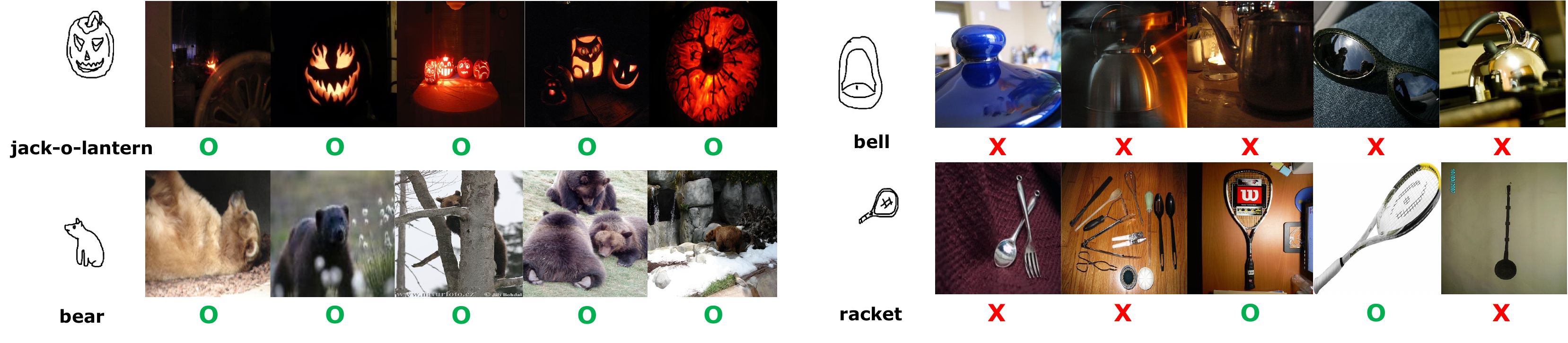}
    \vspace{-1.5em}
    \caption{
    A few examples of retrieved samples from the Sketchy Extended dataset. For each query sketch, the top-5 retrieved images are shown, where the top-1 is in the leftmost.
    The O/X's indicate whether the retrievals belong to the same class of the query.
    }
    \label{fig:retrieval}
\end{figure*}

Some examples of retrieved photos are shown in Fig.~\ref{fig:retrieval}.
Note that even the falsely retrieved photos share visual similarity with the query sketches.

\section{Concluding Remarks}\label{sec:concluding}
Cuff's channel synthesis and Wyner's distributed simulation provide an information-theoretic characterization of the simplest probabilistic structure that connects one random object to another.
The proposed variational Wyner model finds this succinct structure in a disciplined manner, and provides a theoretically sound alternative to the information bottleneck principle~\citep{Tishby--Pereira--Bialek1999}. 
As alluded to earlier in Section~\ref{sec:related},
our approach is the first to define an optimal common representation and learn a generative model towards the optimality.
The experimental results demonstrated the potential of our approach as a new way of learning joint and conditional generation tasks with optimal representation learning that can be further developed and refined for more complex dataset such as auditory, text, or a pair of those.

Albeit its potentially wide applicability, we remark that the proposed model and the accompanied training method may suffer slow training and memory inefficiency, as each divergence and mutual information term requires a separate density ratio estimator.
While we reduce the number of parameters by sharing a joint feature map in the discriminator, it might be crucial to devise a more efficient way to implement the proposed framework with less parameters.

We conclude with future directions.
First, we assumed fully paired data throughout the paper. In practice, however, paired data are limited and we have a plenty of unpaired data. 
Investigating on how to incorporate such unpaired data in the current learning framework and studying the role and effect of common information regularization in the \emph{semi-supervised} setting may be a fruitful direction, which will make the developed framework applicable in a much richer context.
Second, in this paper, we motivated the use of  Wyner's common information via a heuristic justification from the resemblance between the generative tasks and the information theory problems. Hence, it would be also interesting to formally establish an operational meaning of the common information $I(\Xv,\Yv;\Zv)$ when learning distributions from samples. For example, can we develop a theory that relates the common information $I(\Xv,\Yv;\Zv)$ with its ``generalization error'' for a generative model as in~\cite{Xu--Raginsky2017}?

\section*{Acknowledgment}
This work was supported in part by the National Science Foundation under Grant CCF-1911238 and SoC R\&D, Samsung Semiconductor, Inc.

\ifCLASSOPTIONcaptionsoff
  \newpage
\fi

\appendices
\section{From Minimal Sufficient Statistics to the Information Bottleneck Principle and Wyner's Optimization Problem}
\label{app:minimal}

In this section, starting from the notion of minimal sufficient statistics~\cite{Lehmann--Scheffe1950} from the statistics literature, we derive the information bottleneck (IB) principle and Wyner's optimization problem from its relaxation, respectively.
We hope that this discussion highlights similarities and dissimilarities between the IB principle and Wyner's optimization problem.

\subsection{Minimal Sufficient Statistics}
Consider a pair of random variables $(\Xv,\Yv)$.
A function $\Zv=\zv(\Xv)$ of $\Xv$ is said to be a \emph{minimal sufficient statistic} of $\Xv$ for $\Yv$ if (1) (sufficiency) $\Zv$ is a sufficient statistic of $\Xv$ for $\Yv$, \ie $\Xv\leftrightarrow\Zv\leftrightarrow\Yv$, and (2) (minimality) $\Zv$ is a function of any other sufficient statistics.
It can be easily shown that $\Zv=\zv(\Xv)$ is a minimal sufficient statistic if and only if it is an optimal solution of the following optimization problem
\begin{align}
\begin{aligned}
& \minimize && I(\Xv;\Zv)\\
& \text{~subject to} &&
    \Xv\leftrightarrow\Zv\leftrightarrow\Yv\\
& \text{~variable} &&
    \Zv=\zv(\Xv).
\end{aligned}
\label{eq:minimal_sufficient_statistic}
\end{align}
Here, the optimization is over all possible functions $\zv(\cdot)$ over $\xv$.

\subsection{The IB Principle}\label{app:ib}
First, note that the Markovity constraint $\Xv\leftrightarrow\Zv\leftrightarrow\Yv$ in \eqref{eq:minimal_sufficient_statistic} can be relaxed as $I(\Yv;\Zv)\ge I(\Xv;\Yv)$ by the data processing inequality.
Second, optimization over functions $\zv(\xv)$ can be relaxed by considering optimization over a probabilistic mapping $q(\zv|\xv)$, as it subsumes deterministic functions.
Finally, introducing a Lagrangian multiplier $\b>0$ to get rid of the inequality constraint on $I(\Yv;\Zv)$, we obtain a relaxed version of \eqref{eq:minimal_sufficient_statistic} in the unconstrained form
\begin{align}
\begin{aligned}
& \minimize && I(\Xv;\Zv)-\b I(\Yv;\Zv)\\
& \text{~variable} &&
    q(\zv|\xv).
\end{aligned}
\label{eq:ib}
\end{align}
Note that this is the optimization problem that characterizes the IB principle.
A similar argument can be found in \citep{Shamir--Sabato--Tishby2010}.

\subsection{Wyner's Optimization Problem}
As done above, we first relax optimization over $\zv(\xv)$ by optimization over $q(\zv|\xv)$.
Then, observe that optimizing over $q(\zv|\xv)$ is equivalent to $q(\zv|\xv,\yv)$ under an additional constraint $\Zv\leftrightarrow\Xv\leftrightarrow\Yv$.
Hence, we can relax the optimization problem~\eqref{eq:minimal_sufficient_statistic} as
\begin{align}
\begin{aligned}
& \minimize && I(\Xv;\Zv)\\
& \text{~subject to} &&
    \Xv\leftrightarrow\Zv\leftrightarrow\Yv\\
& && \Zv\leftrightarrow\Xv\leftrightarrow\Yv\\
& \text{~variable} &&
    q(\zv|\xv,\yv).
\end{aligned}
\label{eq:wyner_intermediate}
\end{align}
By removing the Markovity constraint $\Zv\leftrightarrow\Xv\leftrightarrow\Yv$, we can further relax it as
\begin{align}
\begin{aligned}
& \minimize && I(\Xv;\Zv)\\
& \text{~subject to} &&
    \Xv\leftrightarrow\Zv\leftrightarrow\Yv\\
& \text{~variable} &&
    q(\zv|\xv,\yv).
\end{aligned}
\label{eq:wyner_intermediate2}
\end{align}
We claim that this is equivalent to Wyner's optimization problem~\eqref{eq:wyner}.
Indeed, note that we have $I(\Xv,\Yv;\Zv)=I(\Xv;\Zv)+h(\Yv|\Xv)$ under the Markov chain $\Xv\to\Zv\to\Yv$, since
\begin{align*}
I(\Xv,\Yv;\Zv)
&=I(\Xv;\Zv)+I(\Yv;\Zv|\Xv)\\
&=I(\Xv;\Zv)+h(\Yv|\Xv)+h(\Yv|\Xv,\Zv)\\
&=I(\Xv;\Zv)+h(\Yv|\Xv).
\end{align*}
Here, $h(\Yv|\Xv,\Zv)=0$ follows from the Markov chain $\Xv\leftrightarrow\Zv\leftrightarrow\Yv$.
Since $h(\Yv|\Xv)$ is constant given the target distribution $q(\xv,\yv)$,
\eqref{eq:wyner_intermediate2} is equivalent to \eqref{eq:wyner}.

\subsection{Discussion}
We remark that we can derive an optimization problem from \eqref{eq:wyner_intermediate2} that is directly comparable to the IB principle~\eqref{eq:ib}.
By applying the same argument in Appendix~\ref{app:ib} to \eqref{eq:wyner_intermediate2}, we can relax $\Xv\leftrightarrow\Zv\leftrightarrow\Yv$ to $I(\Yv;\Zv)\ge I(\Xv;\Yv)$ and convert it into an unconstrained problem
\begin{align}
\begin{aligned}
& \minimize && I(\Xv;\Zv)-\b I(\Yv;\Zv)\\
& \text{~variable} &&
    q(\zv|\xv,\yv).
\end{aligned}
\label{eq:wyner_relaxed}
\end{align}
Interestingly, this has a close resemblance to the IB principle~\ref{eq:ib}.
In particular, \eqref{eq:wyner_relaxed} can be viewed as a relaxation of \ref{eq:ib}, since optimizing over $q(\zv|\xv,\yv)$ subsumes optimizing over $q(\zv|\xv)$.

\section{Deferred technical statements}
The following proposition justifies the form of local variational encoders $q_\phi(\uv|\zv,\xv)$ and $q_\phi(\vv|\zv,\yv)$, as noted in Remark~\ref{rem:conditional_independence}.
\begin{proposition}
\label{prop:var_encoders}
If $(\Xv,\Yv,\Zv,\Uv,\Vv)\sim p(\zv)p(\uv)p(\vv)p(\xv|\zv,\uv)p(\yv|\zv,\vv)$, 
then
\[
p(\uv,\vv|\zv,\xv,\yv)=p(\uv|\zv,\xv)p(\vv|\zv,\yv),
\]
\ie $\Uv$ and $\Vv$ are conditionally independent given $(\Zv,\Xv,\Yv)$, 
$\Uv$ and $\Yv$ are conditionally independent given $(\Zv,\Xv)$, 
and $\Vv$ and $\Xv$ are conditionally independent given $(\Zv,\Yv)$.
\end{proposition}
\begin{proof}
The conditional independence of the joint distribution $p(\zv)p(\uv)p(\vv)p(\xv|\zv,\uv)p(\yv|\zv,\vv)$ is encoded as the following directed graphical model.
\begin{center}
\includegraphics[width=0.225\textwidth]{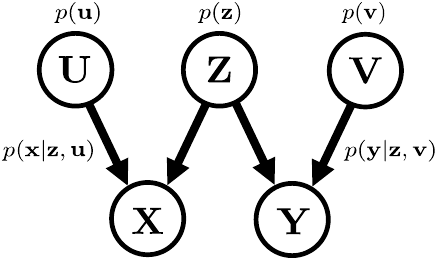}
\end{center}
The desired conditional independences now follow from checking the d-separation~\citep{Koller--Friedman2009} between the nodes. 
That is, $\Uv$ and $\Vv$ are d-separated by $(\Zv,\Xv,\Yv)$, $\Uv$ and $\Yv$ are d-separated by $(\Zv,\Xv)$, and $\Vv$ and $\Xv$ are d-separated by $(\Zv,\Yv)$.
\end{proof}

\begin{proposition}[Monotonicity]
\label{prop:fdiv_monotone}
For a convex function $f\suchthat\Real_+\to\Real$, let $D_f(\cdot\|\cdot)$ denote the $f$-divergence.
For any joint distributions $p_1(x,y)$ and $p_2(x,y)$ over $\Xc\times \Yc$, we have
\[
D_f(p_1(x)\|p_2(x)) \le D_f(p_1(x,y)\|p_2(x,y)).
\]
Here, $p_i(x)$ is the marginal distribution over $\Xc$ induced by $p_i(x,y)$.
In particular, the equality holds if $p_2(y|x)\equiv p_1(y|x)$ for all $x\in\Xc$.
When $f$ is strictly convex, the condition becomes also necessary for the equality.
\end{proposition}
\begin{proof}
Consider
\begin{align*}
&D_f(p_1(x,y)\|p_2(x,y))\\
&=\int p_1(x,y)f\Bigl(\frac{p_2(x,y)}{p_1(x,y)}\Bigr)\diff x \diff y\\
&=\int p_1(x)p_1(y|x)f\Bigl(\frac{p_2(x)p_2(y|x)}{p_1(x)p_1(y|x)}\Bigr)\diff x \diff y\\
&\ge \int p_1(x) f\Bigl(\int p_1(y|x)\frac{p_2(x)p_2(y|x)}{p_1(x)p_1(y|x)}\diff y\Bigr)\diff x\\ 
&= \int p_1(x) f\Bigl( \frac{p_2(x)}{p_1(x)}\Bigr)\diff x
= D_f(p_1(x)\|p_2(x)).
\end{align*}
Here, we use the convexity of $f$ and Jensen's inequality. The equality condition follows from that of the Jensen's inequality.
\end{proof}

\section{Experiment Details}
\label{app:exp_settings}

\subsection{Common Settings}

We used NVIDIA TITAN X (Pascal) for our experiments.
We implemented all models using \href{https://pytorch.org/}{PyTorch}~\citep{Paszke--Gross--Chintala--Chanan--Yang--DeVito--Lin--Desmaison--Antiga--Lerer2017pytorch}.
For the inference phase, we applied the exponential moving average with decay parameter 0.999.
As alluded to earlier, all distributions are parameterized by deterministic models.

\textbf{Notation.}
Let \texttt{fc(c\_in,c\_out)} denote a fully-connected layer with \texttt{c\_in} input units and \texttt{c\_out} output units.
Let \texttt{conv2d(c,k,s,p)} (\texttt{deconv2d(c,k,s,p)}) denote a two-dimensional convolutional (transposed convolutional) layer with $c$ filters, kernel of size $\texttt{k}\times \texttt{k}$, stride $(\texttt{s},\texttt{s})$, and zero-padding of size $(\texttt{p},\texttt{p})$.
Let \texttt{conv2d(c,(k1,k2),(s1,s2),(p1,p2))} (\texttt{deconv2d(c,(k1,k2),(s1,s2),(p1,p2))}) denote a two-dimensional convolutional (transposed convolutional) layer with $\texttt{c}$ filters, kernel of size $\texttt{k1}\times \texttt{k2}$, stride $(\texttt{s1},\texttt{s2})$, and zero-padding of size $(\texttt{p1},\texttt{p2})$.
We use $\texttt{dropout1d}$ and $\texttt{dropout2d}$ to denote 1D and 2D dropout layers, and use $\texttt{maxpool2d(k)}$ to denote a 2D max pooling layer with kernel of size (\texttt{k},\texttt{k}).

\subsection{MNIST--SVHN Add-One}
In this dataset, $\xv=\text{MNIST}$, $\yv=\text{SVHN}$.
For the MNIST--SVHN add-one dataset, we constructed $50,000$ add-one pairs from the MNIST and SVHN training datasets. 
For testing, we similarly constructed $1,000$ paired images from MNIST and SVHN test datasets in each case. 
We padded zeros around the $28\times 28$ MNIST images to make them of size $32\times 32$.
All pixel values were linearly translated to range between $[-1,1]$.

\subsubsection{Evaluation Metrics}

\begin{table}[t]
\centering
\caption{The neural network architecture of the symmetric decoder in the MNIST and SVHN autoencoders. We used 1 and 3 for \texttt{c\_out}, respectively for MNIST and SVHN datasets. This architecture was used to evaluate the Frechet distance; see Section~\ref{sec:fd_score}.}
\label{tab:ae_decoder}
\begin{tabular}[t]{c}
\toprule
$\psi_{\textsf{feature}\to\textsf{image}}$ \\
\midrule
\texttt{deconv2d(128,5,2,2)-bn2d-LReLU(0.2)}\\
\texttt{deconv2d(64,5,2,2)-bn2d-LReLU(0.2)}\\
\texttt{deconv2d(32,5,2,2)-bn2d-LReLU(0.2)}\\
\texttt{deconv2d(\texttt{c\_out},5,2,2)}\\
\bottomrule
\end{tabular}
\end{table}

\paragraph{Frechet distance}
\label{sec:fd_score}
To compute the Frechet distance (FD) score for the joint and conditional distributions over (MNIST, SVHN) pairs, we implemented a customized Frechet distance based on the PyTorch implementation of the Frechet Inception distance (FID) score~\citep{Heusel--Ramsauer--Unterthiner--Nessler--Hochreiter2017TTUR} by \citet{Seitzer2020FID}\footnote{\url{https://github.com/mseitzer/pytorch-fid}}.
Essentially, to be better tailored to the digit images of MNIST and SVHN datasets, we replaced the Inception-v3 model with pretrained feature extractors trained from autoencoders for MNIST and SVHN, respectively.
We used the network architectures defined in Tables~\ref{tab:ae_decoder} and \ref{tab:architectures_mnist_svhn_encoders} and defined autoencoders  \[\xv_\texttt{image}\mapsto \psi_{\textsf{feature$\to$image}}^{\texttt{image}}((f_{\textsf{image$\to$feature}}^{\texttt{image}}(\xv_\texttt{image}))\] for \texttt{image$\in$\{mnist,svhn\}}.
We trained the MNIST and SVHN autoencoders for 200 and 25 epochs, respectively, with Adam optimizer with learning rate $10^{-4}$ and batch size 64.
Note that we used the ``extra'' split of the SVHN dataset for training.
After training, we used the encoders as feature extractors.
To evaluate the joint distribution, we concatenated the two feature vectors to compute the mean and covariance for each dataset.
To evaluate the conditional distribution, we computed the FD score for each digit class separately and reported the averaged values over the 10 classes.
Note that the FD score was evaluated with respect to the test datasets of MNIST and SVHN, so that a low FD score implies that the model generates similar images to (unseen) test images.

\paragraph{Digit classification error}
For the digit classification error reported in Fig.~\ref{fig:mnist_svhn_numerical}, we used pretrained classifiers with network architectures in Table~\ref{tab:mnist_svhn_classifier}. Each classifier was trained for 15 epochs with the cross entropy loss and Adam optimizer with learning rate $10^{-3}$ and batch size 32. 
To evaluate a conditional accuracy, we computed an accuracy for each class and reported their average.

\begin{table}[t]
\centering
\caption{The neural network architecture of the MNIST and SVHN classifiers. Note that we used the identical architecture, and it only differs in the bottleneck dimension due to the difference in the numbers of channels.}
\label{tab:mnist_svhn_classifier}
\begin{tabular}{c}
\toprule
MNIST classifier\\
\midrule
\texttt{conv2d(10,5,1,0)-maxpool2d(2)-ReLU}\\
\texttt{conv2d(20,5,1,0)-dropout2d-maxpool2d(2)-ReLU}\\
\texttt{reshape(batch\_size, 320)}\\
\texttt{fc(320, 50)-maxpool2d(2)-ReLU-dropout1d}\\
\texttt{fc(50, 10)-softmax}\\
\bottomrule
\end{tabular}\vspace{1em}

\begin{tabular}{c}
\toprule
SVHN classifier\\
\midrule
\texttt{conv2d(10,5,1,0)-maxpool2d(2)-ReLU}\\
\texttt{conv2d(20,5,1,0)-dropout2d-maxpool2d(2)-ReLU}\\
\texttt{reshape(batch\_size, 500)}\\
\texttt{fc(500, 50)-maxpool2d(2)-ReLU-dropout1d}\\
\texttt{fc(50, 10)-softmax}\\
\bottomrule
\end{tabular}
\end{table}

\subsubsection{Network Architectures}
The neural network architectures are summarized in Tables~\ref{tab:architectures_mnist_svhn_encoders}--\ref{tab:architectures_mnist_svhn_discriminators}; here, $f$'s are used in encoders, $g$'s are in decoders, and $h$'s are in discriminators.

Define $f_{\textsf{aggregate}}^{\texttt{joint}}(\xv,\yv)\defeq f_{\textsf{image$\to$feature}}^{\texttt{joint}}(\xv,\yv)$.

\begin{table}[th]
\centering
\caption{The neural network architectures in the MNIST--SVHN encoders.\\ 
The output of the image feature network $f_{\textsf{image$\to$feature}}(\xv)$ has dimension (batch\_size, 1024).}
\label{tab:architectures_mnist_svhn_encoders}
\scriptsize
\begin{tabular}[t]{c}
\toprule
$f_{\textsf{image$\to$feature}}$ \\
\midrule
\texttt{conv2d(32,5,2,2)-bn2d-LReLU(0.2)}\\

\texttt{conv2d(64,5,2,2)-bn2d-LReLU(0.2)}\\

\texttt{conv2d(128,5,2,2)-bn2d-LReLU(0.2)}\\

\texttt{conv2d(256,5,2,2)-bn2d-LReLU(0.2)}\\

\texttt{flatten}\\
\bottomrule
\end{tabular}
\quad
\begin{tabular}[t]{c}
\toprule
$f_{\textsf{feature$\to$\zv}}$ \\
\midrule
\texttt{fc(1024,dim\_z)}\\
\bottomrule
\end{tabular}\vspace{1em}

\begin{tabular}[t]{c}
\toprule
$f_{\textsf{$\zv\to$latent\_feature}}$ \\
\midrule
\texttt{fc(dim\_z,1024)-bn1d-LReLU(0.2)}\\
\bottomrule
\end{tabular}
\quad
\begin{tabular}[t]{c}
\toprule
$f_{\textsf{features$\to$\uv}}$ \\
\midrule
\texttt{fc(2048,dim\_local)}\\
\bottomrule
\end{tabular}
\end{table}%

\begin{table}[th]
\centering
\caption{The neural network architectures in the MNIST--SVHN decoders.}
\label{tab:architectures_mnist_svhn_decoders}
\scriptsize
\begin{tabular}[t]{c}
\toprule
$g_{(\zv,\uv)\to\textsf{feature}}$ \\
\midrule
\texttt{fc(dim\_z+dim\_u,8192)-bn1d-LReLU(0.2)}\\
\texttt{reshape(batch\_size,512,4,4)}\\
\bottomrule
\end{tabular}\vspace{1em}

\begin{tabular}[t]{c}
\toprule
$g_{\textsf{feature$\to$image}}$ \\
\midrule
\texttt{deconv2d(256,5,2,2)-bn2d-LReLU(0.2)}\\
\texttt{deconv2d(128,5,2,2)-bn2d-LReLU(0.2)}\\
\texttt{deconv2d(c\_out,5,2,2)-bn2d-LReLU(0.2)}\\
\bottomrule
\end{tabular}
\end{table} %

\begin{table}[th]
\centering
\caption{The neural network architectures in the MNIST--SVHN discriminators.\\
The output of the image feature network $h_{\textsf{image$\to$feature}}(\xv)$ has dimension (batch\_size, 2048).}
\label{tab:architectures_mnist_svhn_discriminators}
\scriptsize
\begin{tabular}[t]{c}
\toprule
$h_{\textsf{image$\to$feature}}^{\texttt{common}}$ \\
\midrule
\texttt{conv2d(64,5,2,2)-LReLU(0.2)}\\
\texttt{conv2d(128,5,2,2)-bn2d-LReLU(0.2)}\\
\texttt{conv2d(256,5,2,2)-bn2d-LReLU(0.2)}\\
\texttt{conv2d(512,5,2,2)-bn2d-LReLU(0.2)}\\
\texttt{flatten}\\
\bottomrule
\end{tabular}\vspace{1em}

\begin{tabular}[t]{c}
\toprule
$h_{\textsf{latent}\to\textsf{feature}}$ \\
\midrule
\texttt{fc(total\_latent\_dim,512)-LReLU(0.2)}\\
\bottomrule
\end{tabular}\vspace{1em}

\begin{tabular}[t]{c}
\toprule
$h_{\textsf{feature$\to$ratio}}$ \\
\midrule
\texttt{fc(1536,512)-bn1d-LReLU(0.2)}\\
\texttt{fc(512,1)}\\
\bottomrule
\end{tabular}
\end{table}

\paragraph{Generator Models}

\begin{itemize}
\item MNIST encoder / decoder
\begin{itemize}
\item $\zv\sim q(\zv|\xv)$\\
$ \equiv \zv\gets f_{\textsf{feature}\to\zb}^{\texttt{mnist}}(f_{\textsf{image$\to$feature}}^{\texttt{mnist}}(\xv))$.
\item $\uv\sim q(\uv|\zv,\xv)$\\ $\equiv \uv \gets f_{\textsf{features$\to$\uv}}^{\texttt{mnist}}(f_{\zv\to\textsf{latent\_feature}}^{\texttt{mnist}}(\zv),\xv)$.
\item $\xv\sim p(\xv|\zv,\uv)$ \\
$\equiv \xv \gets g_{\textsf{feature$\to$ image}}^{\texttt{mnist}}(g_{(\zv,\uv)\to \textsf{feature}}^{\texttt{mnist}}(\zv,\uv))$.
\end{itemize}

\item SVHN encoder / decoder
\begin{itemize}
\item $\zv\sim q(\zv|\yv)$\\
$\equiv \zv\gets f_{\textsf{feature}\to\zb}^{\texttt{svhn}}(f_{\textsf{image$\to$feature}}^{\texttt{svhn}}(\yv))$.
\item $\vv\sim q(\vv|\zv,\yv)$\\
$ \equiv \vv \gets
f_{\textsf{features$\to$\uv}}^{\texttt{svhn}}(f_{\zv\to\textsf{latent\_feature}}^{\texttt{svhn}}(\zv),\yv)$.
\item $\yv\sim p(\yv|\zv,\vv)$\\
$ \equiv \yv \gets g_{\textsf{feature$\to$ image}}^{\texttt{svhn}}(g_{(\zv,\vv)\to \textsf{feature}}^{\texttt{svhn}}(\zv,\vv))$.
\end{itemize}

\item (MNIST, SVHN) encoder
\begin{itemize}
    \item $\zv\sim q(\zv|\xv,\yv) $\\
$\equiv \zv\gets f_{\textsf{feature}\to\zb}^{\texttt{joint}}(f_{\textsf{aggregate}}^{\texttt{joint}}(\xv,\yv))$.
\end{itemize}
\end{itemize}

\paragraph{Discriminator Models}
As noted in Section~\ref{sec:training}, all the discriminators shared the same feature network, \ie $h_{\textsf{image}\to\textsf{feature}}^{\texttt{common}}(\xv,\yv)$.
\begin{itemize}
    \item Each discriminator to match distributions for $\texttt{model}\in\{\joint,\xtoy,\ytox\}$ (see \eqref{eq:obj_matching_model_xyzuv} and/or \eqref{eq:obj_matching_cross_xyzuv} for the model loss and \eg \eqref{eq:disc_obj_matching_toxy} for the discriminator loss) has the following form:
    \begin{align*}
    &r^\texttt{model}(\xv,\yv,\zv,\uv,\vv)\\
    &=h_{\textsf{feature}\to\textsf{ratio}}^\texttt{model}(h_{\textsf{image}\to\textsf{feature}}^\texttt{common}(\xv,\yv), h_{\textsf{latent}\to\textsf{feature}}^\texttt{model}(\zv,\uv,\vv)).
    \end{align*}
    \item Each discriminator to compute common information for $\texttt{model}\in\{\joint,\xtoy,\ytox\}$~(see \eqref{eq:obj_ci_model} for the model loss and \eqref{eq:disc_obj_ci_model} for the discriminator loss) has the following form: 
    \begin{align*}
    &r^\texttt{model,ci}(\xv,\yv,\zv)\\
    &=h_{\textsf{feature}\to\textsf{ratio}}^\texttt{model,ci}(h_{\textsf{image}\to\textsf{feature}}^\texttt{common}(\xv,\yv), h_{\textsf{latent}\to\textsf{feature}}^\texttt{model,ci}(\zv)).
    \end{align*}
    \item Each discriminator to compute the latent matching loss for $\texttt{model}\in\{\joint,\xtoy,\ytox\}$~(see \eqref{eq:obj_matching_z} for the model loss and \eqref{eq:disc_obj_matching_latent} for the discriminator loss) has the following form:
    \begin{align*}
    r^\texttt{model,agg}(\zv)
    =h_{\textsf{feature}\to\textsf{ratio}}^\texttt{model,agg}(h_{\textsf{latent}\to\textsf{feature}}^\texttt{model,agg}(\zv)).
    \end{align*}
\end{itemize}

\subsubsection{Training}
We used the Adam optimizer~\citep{Kingma--Ba2014} with $(\b_1,\b_2)=(0.5,0.999)$, learning rate $10^{-4}$ for the generators and $2\times 10^{-4}$ for the discriminators~\citep{Heusel--Ramsauer--Unterthiner--Nessler--Hochreiter2017TTUR}.
We trained for 25 epochs.

In the discriminators, we added a mean-zero gaussian noise of standard deviation 0.25 for each variable $\xv,\yv,\zv,\uv,\vv$.

\subsection{CUB Image--Caption}
In this dataset, $\xv=\text{Image (pretrained ResNet features)}$, $\yv=\text{Caption}$.

\subsubsection{Evaluation Metrics}
We followed the same evaluation procedure of \citep{Massiceti--Dokania--Siddharth--Torr2018} and \citep[Section~4.3]{Shi--Siddharth--Paige--Torr2019MMVAE}, which is to perform the canonical correlation analysis (CCA)~\citep{Hotelling1936} and report the correlation score with respect to feature vectors of image and caption.
For the image dataset, recall that we already have 2048-dim. features from the pretrained ResNet-101, which were used in training.
For each caption, we trained a FastText model~\citep{Bojanowski--Grave--Joulin--Mikolog2017} using all sentences in the training dataset to convert each word to a 300-dim. vector; an embedding of a caption was obtained by taking the average embedding of each word in the sentence.
After extracting features, we performed the CCA with projection dimension 40.
For the jointly generated samples, we used 1000 samples.
For the conditionally generated samples, we used the entire test set and reported the average score.

\subsubsection{Network Architectures}
\label{app:network_architectures_mnist_svhn}
The following architectures were adopted from \citep{Shi--Siddharth--Paige--Torr2019MMVAE} with some adjustments.
When processing the caption, the maximum sentence length was 32 and the embedding dimension was 128.
The neural network architectures are summarized in Tables~\ref{tab:architectures_cub_encoders}--\ref{tab:architectures_cub_discriminators}; 
here, $f$'s are used in encoders, $g$'s are in decoders, and $h$'s are in discriminators.
Again, we note that a network without superscript indicates that the same network architecture is used in multiple places with different (\ie not shared) realizations.

We made the embedding layer trainable as well.

\paragraph{Generator Models}
\begin{itemize}
\item Image encoder / decoder
\begin{itemize}
\item $\zv\sim q(\zv|\xv)$\\
$\equiv \zv\gets f_{\textsf{feature}\to\zb}^{\texttt{image}}(f_{\textsf{image$\to$feature}}^{\texttt{image}}(\xv))$.
\item $\uv\sim q(\uv|\zv,\xv)$\\
$ \equiv \uv \gets f_{\textsf{features$\to$\uv}}^{\texttt{image}}(f_{\zv\to\textsf{latent\_feature}}^{\texttt{image}}(\zv),\xv)$.
\item $\xv\sim p(\xv|\zv,\uv)$\\
$ \equiv \xv \gets g_{\textsf{feature$\to$ image}}^{\texttt{image}}(g_{(\zv,\uv)\to \textsf{feature}}^{\texttt{image}}(\zv,\uv))$.
\end{itemize}

\item Caption encoder / decoder
\begin{itemize}
\item $\zv\sim q(\zv|\yv)$\\
$ \equiv \zv\gets f_{\textsf{feature}\to\zb}^{\texttt{sent}}(f_{\textsf{sent$\to$feature}}^{\texttt{sent}}(\yv))$.
\item $\vv\sim q(\vv|\zv,\yv)$\\
$ \equiv \vv \gets f_{\textsf{(\zv,feature)$\to$\vv}}^{\texttt{sent}}(f_{\zv\to\textsf{latent\_feature}}^{\texttt{sent}}(\zv),\yv)$.
\item $\yv\sim p(\yv|\zv,\vv)$\\
$ \equiv \yv \gets g_{\textsf{feature$\to$ image}}^{\texttt{sent}}(g_{(\zv,\vv)\to \textsf{feature}}^{\texttt{sent}}(\zv,\vv))$.
\end{itemize}
\item (Image, Caption) encoder
\begin{itemize}
    \item $\zv\sim q(\zv|\xv,\yv)$\\
$ \equiv \zv\gets f_{\textsf{feature}\to\zb}^{\texttt{joint}}(f_{\textsf{aggregate}}^{\texttt{joint}}(\xv,\yv))$.
\end{itemize}
\end{itemize}

\paragraph{Discriminator Models} 
As noted in Section~\ref{sec:training}, all the discriminators shared the same feature network, \ie $h_{\textsf{(image,sent)}\to\textsf{feature}}^{\texttt{common}}(\xv,\yv)\defeq h_{\textsf{aggregate}}^{\texttt{common}}(h_{\textsf{image}\to\textsf{feature}}^{\texttt{common}}(\xv),h_{\textsf{sent}\to\textsf{feature}}^{\texttt{common}}(\yv))$. Note that the following definitions are almost equivalent to the discriminators for the MNIST--SVHN model except the form of the shared joint feature map.
\begin{itemize}
    \item Each discriminator to match distributions for $\texttt{model}\in\{\joint,\xtoy,\ytox\}$ (see \eqref{eq:obj_matching_model_xyzuv} and/or \eqref{eq:obj_matching_cross_xyzuv} for the model loss and \eg \eqref{eq:disc_obj_matching_toxy} for the discriminator loss) has the following form:
    \begin{align*}
    &r^\texttt{model}(\xv,\yv,\zv,\uv,\vv)\\
    &=h_{\textsf{feature}\to\textsf{ratio}}^\texttt{model}(h_{\textsf{(image,sent)}\to\textsf{feature}}^{\texttt{common}}(\xv,\yv), h_{\textsf{latent}\to\textsf{feature}}^\texttt{model}(\zv,\uv,\vv)).
    \end{align*}
    \item Each discriminator to compute common information for $\texttt{model}\in\{\joint,\xtoy,\ytox\}$~(see \eqref{eq:obj_ci_model} for the model loss and \eqref{eq:disc_obj_ci_model} for the discriminator loss) has the following form: 
    \begin{align*}
    &r^\texttt{model,ci}(\xv,\yv,\zv)\\
    &=h_{\textsf{feature}\to\textsf{ratio}}^\texttt{model,ci}(h_{\textsf{(image,sent)}\to\textsf{feature}}^{\texttt{common}}(\xv,\yv), h_{\textsf{latent}\to\textsf{feature}}^\texttt{model,ci}(\zv)).
    \end{align*}
    \item Each discriminator to compute the latent matching loss for $\texttt{model}\in\{\joint,\xtoy,\ytox\}$~(see \eqref{eq:obj_matching_z} for the model loss and \eqref{eq:disc_obj_matching_latent} for the discriminator loss) has the following form:
    \begin{align*}
    r^\texttt{model,agg}(\zv)
    &=h_{\textsf{feature}\to\textsf{ratio}}^\texttt{model,agg}(h_{\textsf{latent}\to\textsf{feature}}^\texttt{model,agg}(\zv)).
    \end{align*}
\end{itemize}

\begin{table}[th]
\centering
\caption{The neural network architectures in the CUB encoders.}
\label{tab:architectures_cub_encoders}
\scriptsize
\begin{tabular}[t]{c}
\toprule
$f_{\textsf{image$\to$feature}}$ \\
\midrule
\texttt{fc(2048,1024)-bn1d-LReLU(0.2)}\\
\texttt{fc(1024,512)-bn1d-LReLU(0.2)}\\
\texttt{fc(512,256)-bn1d-LReLU(0.2)}\\
\bottomrule
\end{tabular}
\quad
\begin{tabular}[t]{c}
\toprule
$f_{\textsf{feature$\to\zv$}}^{\texttt{image}}$ \\
\midrule
\texttt{fc(256,dim\_z)}\\
\bottomrule
\end{tabular}\vspace{1em}

\begin{tabular}[t]{c}
\toprule
$f_{\textsf{$\zv\to$latent\_feature}}^{\texttt{image}}$ \\
\midrule
\texttt{fc(dim\_z,256)-bn1d-LReLU(0.2)}\\
\bottomrule
\end{tabular}
\quad
\begin{tabular}[t]{c}
\toprule
$f_{\textsf{features$\to$\uv}}^{\texttt{image}}$ \\
\midrule
\texttt{fc(512,dim\_u)}\\
\bottomrule
\end{tabular}\vspace{1em}

\begin{tabular}[t]{c}
\toprule
$f_{\textsf{sent$\to$feature}}$ \\
\midrule
\texttt{embedding(1590,128)}\\
\texttt{reshape(batch\_size,128,32)}\\
\texttt{conv2d(32,4,2,1)-bn2d-LReLU(0.2)}\\
\texttt{conv2d(64,4,2,1)-bn2d-LReLU(0.2)}\\
\texttt{conv2d(128,4,2,1)-bn2d-LReLU(0.2)}\\
\texttt{conv2d(256,(1,4),(1,2),(0,1))-bn2d-LReLU(0.2)}\\
\texttt{conv2d(512,(1,4),(1,2),(0,1))-bn2d-LReLU(0.2)}\\
\bottomrule
\end{tabular}\vspace{1em}

\begin{tabular}[t]{c}
\toprule
$f_{\textsf{feature$\to\zv$}}^{\texttt{sent}}$ \\
\midrule
\texttt{fc(8192,dim\_z)}\\
\bottomrule
\end{tabular}
\quad
\begin{tabular}[t]{c}
\toprule
$f_{\textsf{$\zv\to$latent\_feature}}^{\texttt{sent}}$ \\
\midrule
\texttt{fc(dim\_z,8192)-bn1d-LReLU(0.2)}\\
\bottomrule
\end{tabular}\vspace{1em}

\begin{tabular}[t]{c}
\toprule
$f_{\textsf{features$\to$\vv}}^{\texttt{sent}}$ \\
\midrule
\texttt{fc(16384,dim\_v)}\\
\bottomrule
\end{tabular}\vspace{1em}

\begin{tabular}[t]{c}
\toprule
$f_{\textsf{aggregate}}^{\texttt{(image,sent)}}$ \\
\midrule
\texttt{fc(8448,1024)-bn1d-LReLU(0.2)}\\
\texttt{fc(1024,1024)-bn1d-LReLU(0.2)}\\
\texttt{fc(1024,512)-bn1d-LReLU(0.2)}\\
\bottomrule
\end{tabular}
\quad
\begin{tabular}[t]{c}
\toprule
$f_{\textsf{feature$\to$\zv}}^{\texttt{(image,sent)}}$ \\
\midrule
\texttt{fc(512,dim\_z)}\\
\bottomrule
\end{tabular}
\end{table}

\begin{table}[th]
\centering
\caption{The neural network architectures in the CUB decoders.}
\label{tab:architectures_cub_decoders}
\scriptsize
\begin{tabular}[t]{c}
\toprule
$g_{(\zv,\uv)\to\textsf{feature}}^{\texttt{image}}$ \\
\midrule
\texttt{fc(dim\_z+dim\_u,256)-LReLU(0.2)}\\
\bottomrule
\end{tabular}\vspace{1em}

\begin{tabular}[t]{c}
\toprule
$g_{\textsf{feature$\to$image}}^{\texttt{image}}$ \\
\midrule
\texttt{fc(256,512)-bn1d-LReLU(0.2)}\\
\texttt{fc(512,1024)-bn1d-LReLU(0.2)}\\
\texttt{fc(1024,2048)}\\
\bottomrule
\end{tabular}\vspace{1em}

\begin{tabular}[t]{c}
\toprule
$g_{(\zv,\vv)\to\textsf{feature}}^{\texttt{sent}}$\\
\midrule
\texttt{fc(dim\_z+dim\_v,8192)}\\
\texttt{reshape(batch\_size,512,4,4)}\\
\texttt{bn2d-LReLU(0.2)}\\
\bottomrule
\end{tabular}\vspace{1em}

\begin{tabular}[t]{c}
\toprule
$g_{\textsf{feature$\to$sent}}^{\texttt{sent}}$ \\
\midrule
\texttt{reshape(batch\_size,512,4,4)}\\
\texttt{deconv2d(256,(1,4),(1,2),(0,1))-bn2d-LReLU(0.2)}\\
\texttt{deconv2d(128,(1,4),(1,2),(0,1))-bn2d-LReLU(0.2)}\\
\texttt{deconv2d(64,4,2,1)-bn2d-LReLU(0.2)}\\
\texttt{deconv2d(32,4,2,1)-bn2d-LReLU(0.2)}\\
\texttt{deconv2d(1,4,2,1)}\\
\texttt{reshape(batch\_size,32,128)}\\
\texttt{fc(128,1590)}\\
\bottomrule
\end{tabular}
\end{table}

\begin{table}[th]
\centering
\caption{The neural network architectures in the CUB discriminators.}
\label{tab:architectures_cub_discriminators}
\scriptsize
\begin{tabular}[t]{c}
\toprule
$h_{\textsf{image$\to$feature}}^{\texttt{common}}$ \\
\midrule
\texttt{fc(2048,2048)-LReLU(0.2)}\\
\texttt{fc(2048,1024)-bn1d-LReLU(0.2)}\\
\texttt{fc(1024,512)-bn1d-LReLU(0.2)}\\
\bottomrule
\end{tabular}\vspace{1em}

\begin{tabular}[t]{c}
\toprule
$h_{\textsf{sent$\to$feature}}^{\texttt{common}}$ \\
\midrule
\texttt{embedding(1590,128)}\\
\texttt{reshape(batch\_size,128,32)}\\
\texttt{conv2d(64,4,2,1)-bn2d-LReLU(0.2)}\\
\texttt{conv2d(128,4,2,1)-bn2d-LReLU(0.2)}\\
\texttt{conv2d(256,4,2,1)-bn2d-LReLU(0.2)}\\
\texttt{conv2d(512,(1,4),(1,2),(0,1))-bn2d-LReLU(0.2)}\\
\texttt{conv2d(1024,(1,4),(1,2),(0,1))-bn2d-LReLU(0.2)}\\
\bottomrule
\end{tabular}\vspace{1em}

\begin{tabular}[t]{c}
\toprule
$h_{\textsf{aggregate}}^{\texttt{common}}$ \\
\midrule
\texttt{fc(16896,2048)-bn1d-LReLU(0.2)}\\
\texttt{fc(2048,2048)-bn1d-LReLU(0.2)}\\
\texttt{fc(2048,1024)-bn1d-LReLU(0.2)}\\
\bottomrule
\end{tabular}\vspace{1em}

\begin{tabular}[t]{c}
\toprule
$h_{\textsf{latent}\to\textsf{feature}}$ \\
\midrule
\texttt{fc(total\_latent\_dim, 512)-LReLU(0.2)}\\
\bottomrule
\end{tabular}\vspace{1em}

\begin{tabular}[t]{c}
\toprule
$h_{\textsf{feature$\to$ratio}}$ \\
\midrule
\texttt{fc(1536, 512)-bn1d-LReLU(0.2)}\\
\texttt{fc(512,1)}\\
\bottomrule
\end{tabular}

\end{table}

\subsubsection{Training}
We used the Adam optimizer~\citep{Kingma--Ba2014} with $(\b_1,\b_2)=(0.5,0.999)$, learning rate $10^{-4}$ for the generators and $2\times 10^{-4}$ for the discriminators~\citep{Heusel--Ramsauer--Unterthiner--Nessler--Hochreiter2017TTUR}.
We trained for 50 epochs.

In the discriminators, we added a zero-mean Gaussian noise of standard deviation 0.25 for each latent variable $\zv,\uv,\vv$.
For the perturbation in the data variables $\xv,\yv$, we did the following, adaptive noise injection based on the standard deviations of each feature dimension.
\begin{itemize}
\item For $\xv$ (images), we injected noise to the ResNet feature of dimension 2048. We precomputed the standard deviation $\sigma_i^{\textsf{resnet}}$ for each dimension $i$.
For each evaluation of a discriminator that takes the ResNet feature as one of its arguments, we injected a zero-mean Gaussian noise of standard deviation $\a_{\textsf{resnet}}\times \sigma_i^{\textsf{resnet}}$ to the dimension $i$, where we used $\a_{\textsf{resnet}}=2$.
\item For $\yv$ (sentences), we injected noise at an embedding level.
We used an embedding layer that maps a word from the vocabulary of size 1590 to a 128-dimensional dense vectors. 
Whenever we computed a discriminator value, we computed the standard deviation $\sigma_i^{\textsf{embed}}$ of the embedding layer for each embedding dimension $i\in\{1,\ldots,128\}$; note here that the standard deviation changed along training as the embedding layer was set to be trainable.
Then, with a scale of $\a_{\textsf{sent}}=0.05$, we added a zero-mean Gaussian noise of standard deviation $\a_{\textsf{embed}}\times \sigma_i^{\textsf{embed}}$ to the embedding dimension $i$.
\end{itemize}

\subsection{ZS-SBIR}
In this dataset, $\xv=\text{Sketch image (pretrained VGG features)}$, $\yv=\text{Photo image (pretrained VGG features)}$.
We followed the same experiment setting of \citep{Hwang--Kim--Hong--Kim2020IIAE}, including
the network architectures.

\subsubsection{Evaluation Metrics}
In this experiment, we evaluated Precision@100 (P@100) and mean average precision (mAP), by translating the Tensorflow implementation of the codebase\footnote{\url{https://github.com/gr8joo/IIAE}} of \citep{Hwang--Kim--Hong--Kim2020IIAE}, which is in turn based on \citep{Yelamarthi--Reddy--Mishra--Mittal2018,Shen--Liu--Shen--Shao2018}.

\subsubsection{Network Architectures}
The neural network architectures are summarized in Tables~\ref{tab:architectures_zssbir_encoders}--\ref{tab:architectures_zssbir_discriminators}; here, $f$'s are used in encoders, $g$'s are in decoders, and $h$'s are in discriminators.
We note that a network without superscript indicates that the same network architecture is used in multiple places with different (\ie not shared) realizations.

The generators and discriminators are defined in the same manner as for the MNIST--SVHN model (see Appendix~\ref{app:network_architectures_mnist_svhn}), and thus omitted here.

\begin{table}[th]
\centering
\caption{The neural network architectures in the ZS-SBIR encoders.}
\label{tab:architectures_zssbir_encoders}
\scriptsize
\begin{tabular}[t]{c}
\toprule
$f_{\textsf{image$\to$feature}}$ \\
\midrule
\texttt{fc(512,512)-LReLU(0.2)}\\
\bottomrule
\end{tabular}
\quad
\begin{tabular}[t]{c}
\toprule
$f_{\textsf{feature$\to$\zv}}$ \\
\midrule
\texttt{fc(512,dim\_z)}\\
\bottomrule
\end{tabular}\vspace{1em}

\begin{tabular}[t]{c}
\toprule
$f_{\textsf{$\zv\to$latent\_feature}}^{\texttt{image}}$ \\
\midrule
\texttt{fc(dim\_z,512)-LReLU(0.2)}\\
\bottomrule
\end{tabular}
\quad
\begin{tabular}[t]{c}
\toprule
$f_{\textsf{features$\to$\uv}}$ \\
\midrule
\texttt{fc(1024,dim\_local)}\\
\bottomrule
\end{tabular}\vspace{1em}

\begin{tabular}[t]{c}
\toprule
$f_{\textsf{aggregate}}^{\texttt{joint}}$ \\
\midrule
\texttt{fc(1024,512)-LReLU(0.2)}\\
\bottomrule
\end{tabular}
\quad
\begin{tabular}[t]{c}
\toprule
$f_{\textsf{feature$\to$\zv}}^{\texttt{joint}}$ \\
\midrule
\texttt{fc(512,dim\_z)}\\
\bottomrule
\end{tabular}
\end{table}

\begin{table}[th]
\centering
\caption{The neural network architectures in the ZS-SBIR decoders.}
\label{tab:architectures_zssbir_decoders}
\scriptsize
\begin{tabular}[t]{c}
\toprule
$g_{(\zv,\uv)\to\textsf{feature}}$ \\
\midrule
\texttt{fc(dim\_z+dim\_u,128)-ReLU}\\
\bottomrule
\end{tabular}
\quad
\begin{tabular}[t]{c}
\toprule
$g_{\textsf{feature$\to$image}}$ \\
\midrule
\texttt{fc(128,512)}\\
\bottomrule
\end{tabular}
\end{table}

\begin{table}[th]
\centering
\caption{The neural network architectures in the ZS-SBIR discriminators.}
\label{tab:architectures_zssbir_discriminators}
\scriptsize
\begin{tabular}[t]{c}
\toprule
$h_{\textsf{image$\to$feature}}^{\texttt{common}}$ \\
\midrule
\texttt{fc(1024,1024)-LReLU(0.2)}\\
\bottomrule
\end{tabular}\vspace{1em}

\begin{tabular}[t]{c}
\toprule
$h_{\textsf{latent}\to\textsf{feature}}$ \\
\midrule
\texttt{fc(total\_latent\_dim, 512)-LReLU(0.2)}\\
\bottomrule
\end{tabular}\vspace{1em}

\begin{tabular}[t]{c}
\toprule
$h_{\textsf{feature$\to$ratio}}$ \\
\midrule
\texttt{fc(1536, 512)-LReLU(0.2)}\\
\texttt{fc(512,1)}\\
\bottomrule
\end{tabular}

\end{table}

\subsubsection{Training}
We used the Adam optimizer~\citep{Kingma--Ba2014} with $(\b_1,\b_2)=(0.5,0.999)$, learning rate $5\times 10^{-4}$ for the generators and $10^{-3}$ for the discriminators~\citep{Heusel--Ramsauer--Unterthiner--Nessler--Hochreiter2017TTUR}.
We trained for 50 epochs.

The noise injection for discriminators was done similarly to the CUB model.
Namely, we added a zero-mean Gaussian noise of standard deviation 0.5 for each latent variable $\zv,\uv,\vv$,
and we added a zero-mean Gaussian noise with adaptive standard deviation to based on the standard deviations of the features as for the CUB-image data, with the multiplicative scale $\a_{\xv}=\a_{\yv}=0.5$.

\bibliographystyle{IEEEtranN}
\bibliography{IEEEabrv,library}

\end{document}